\documentclass[runningheads]{llncs}
\usepackage{graphicx}

\usepackage{tikz}
\usepackage{comment}
\usepackage{amsmath,amssymb,amssymb} 
\usepackage{color}

\usepackage[accsupp]{axessibility}  

\usepackage{booktabs} 
\usepackage{mathrsfs}
\usepackage{cleveref}
\usepackage{multirow}
\usepackage{comment}
\usepackage{changepage}
\usepackage{wrapfig}
\usepackage{colortbl} 
\usepackage{arydshln} 
\usepackage{algorithm}  
\usepackage{algorithmic}  
{
\centering
\begin{minipage}{#2}
\begin{algorithm}[H]}%
{\end{algorithm}
\end{minipage}
}

\usepackage{footmisc}
\DefineFNsymbols{mySymbols}{{\ensuremath\ast}{\ensuremath\dagger}
{\ensuremath\ddagger}{\ensuremath\mathsection}{\ensuremath\mathparagraph}
{\ensuremath\parallel}{**}{\ensuremath{\dagger\dagger}}{\ensuremath{\ddagger\ddagger}}}
\setfnsymbol{mySymbols}

\begin{document}
\pagestyle{headings}
\mainmatter
\def\ECCVSubNumber{2904}  

\title{CoSCL: Cooperation of Small Continual Learners is Stronger than a Big One} 

\titlerunning{CoSCL: Cooperation of Small Continual Learners is Stronger than a Big One}
%


\author{
Liyuan Wang\inst{\rm 1,2,3}\thanks{Liyuan Wang and Xingxing Zhang contributed equally.} \and Xingxing Zhang\inst{\rm 3}$^{\ast}$ \and Qian Li\inst{\rm 1,2} \and Jun Zhu\inst{\rm 3 \dagger} \and Yi Zhong\inst{\rm 1,2}\thanks{Jun Zhu and Yi Zhong are corresponding authors.}}

\authorrunning{L. Wang et al.}
%
\institute{\, \inst{\rm 1} School of Life Sciences, IDG/McGovern Institute for Brain Research, Tsinghua University. \,\, \inst{\rm 2} Tsinghua-Peking Center for Life Sciences. \,\, \inst{\rm 3} Dept. of Comp. Sci. \\ \& Tech., Institute for AI, BNRist Center, THBI Lab, Tsinghua University. \\
\email{\{wly19,xxzhang2020\}@mails.tsinghua.edu.cn, \{liqian8,dcszj,zhongyithu\}@tsinghua.edu.cn}
}


\maketitle

\begin{abstract}
Continual learning requires incremental compatibility with a sequence of tasks. However, the design of model architecture remains an open question: In general, learning all tasks with a shared set of parameters suffers from severe interference between tasks; while learning each task with a dedicated parameter subspace is limited by scalability. In this work, we theoretically analyze the generalization errors for learning plasticity and memory stability in continual learning, which can be uniformly upper-bounded by (1) discrepancy between task distributions, (2) flatness of loss landscape and (3) cover of parameter space. Then, inspired by the robust biological learning system that processes sequential experiences with multiple parallel compartments, we propose Cooperation of Small Continual Learners (CoSCL) as a general strategy for continual learning. Specifically, we present an architecture with a fixed number of narrower sub-networks to learn all incremental tasks in parallel, which can naturally reduce the two errors through improving the three components of the upper bound. To strengthen this advantage, we encourage to cooperate these sub-networks by penalizing the difference of predictions made by their feature representations. With a fixed parameter budget, CoSCL can improve a variety of representative continual learning approaches by a large margin (e.g., up to 10.64\% on CIFAR-100-SC, 9.33\% on CIFAR-100-RS, 11.45\% on CUB-200-2011 and 6.72\% on Tiny-ImageNet) and achieve the new state-of-the-art performance.
Our code is available at https://github.com/lywang3081/CoSCL.

\end{abstract}

\section{Introduction}
The ability to incrementally learn a sequence of tasks is critical for artificial neural networks. Since the training data distribution is typically dynamic and unpredictable, this usually requires a careful trade-off between learning plasticity and memory stability. In general, excessive plasticity in learning new tasks leads to the catastrophic forgetting of old tasks \cite{mccloskey1989catastrophic}, while excessive stability in remembering old tasks limits the learning of new tasks.
Most efforts in continual learning either use a single model to learn all tasks, which has to sacrifice the performance of each task to find a shared solution \cite{ramesh2021model}; or allocate a dedicated parameter subspace for each task to overcome their mutual interference \cite{fernando2017pathnet,serra2018overcoming}, which usually lacks scalability. Recent work observed that a wider network can suffer from less catastrophic forgetting \cite{mirzadeh2021wide}, while different components such as batch normalization, skip connections and pooling layers play various roles \cite{mirzadeh2022architecture}. Thus, how to achieve effective continual learning in terms of model architecture remains an open question. 

In contrast, the robust biological learning system applies multiple compartments (i.e, sub-networks) to process sequential experiences in parallel, and integrates their outputs in a weighted-sum fashion to guide adaptive behaviors \cite{aso2014neuronal,cohn2015coordinated,modi2020drosophila}.
This provides a promising reference for artificial neural networks.


In this work, we first theoretically analyze the generalization errors of learning plasticity and memory stability in continual learning. We identify that both aspects can be uniformly upper-bounded by (1) \emph{discrepancy between task distributions}, (2) \emph{flatness of loss landscape} and (3) \emph{cover of parameter space}. 
Inspired by the biological strategy, we propose a novel method named Cooperation of Small Continual Learners (CoSCL). Specifically, we design an architecture with multiple narrower sub-networks\footnote{In contrast to a single continual learning model with a wide network, we refer to such narrower sub-networks as ``small'' continual learners.} to learn all incremental tasks in parallel, which can naturally alleviate the both errors through improving the three components. To strengthen this advantage, we further encourage the cooperation of sub-networks by penalizing differences in the predictions of their feature representations. 

\begin{wrapfigure}{r}{0.56\textwidth}
     \vspace{-0.7cm}
	\centering
	\includegraphics[width=0.56\columnwidth]{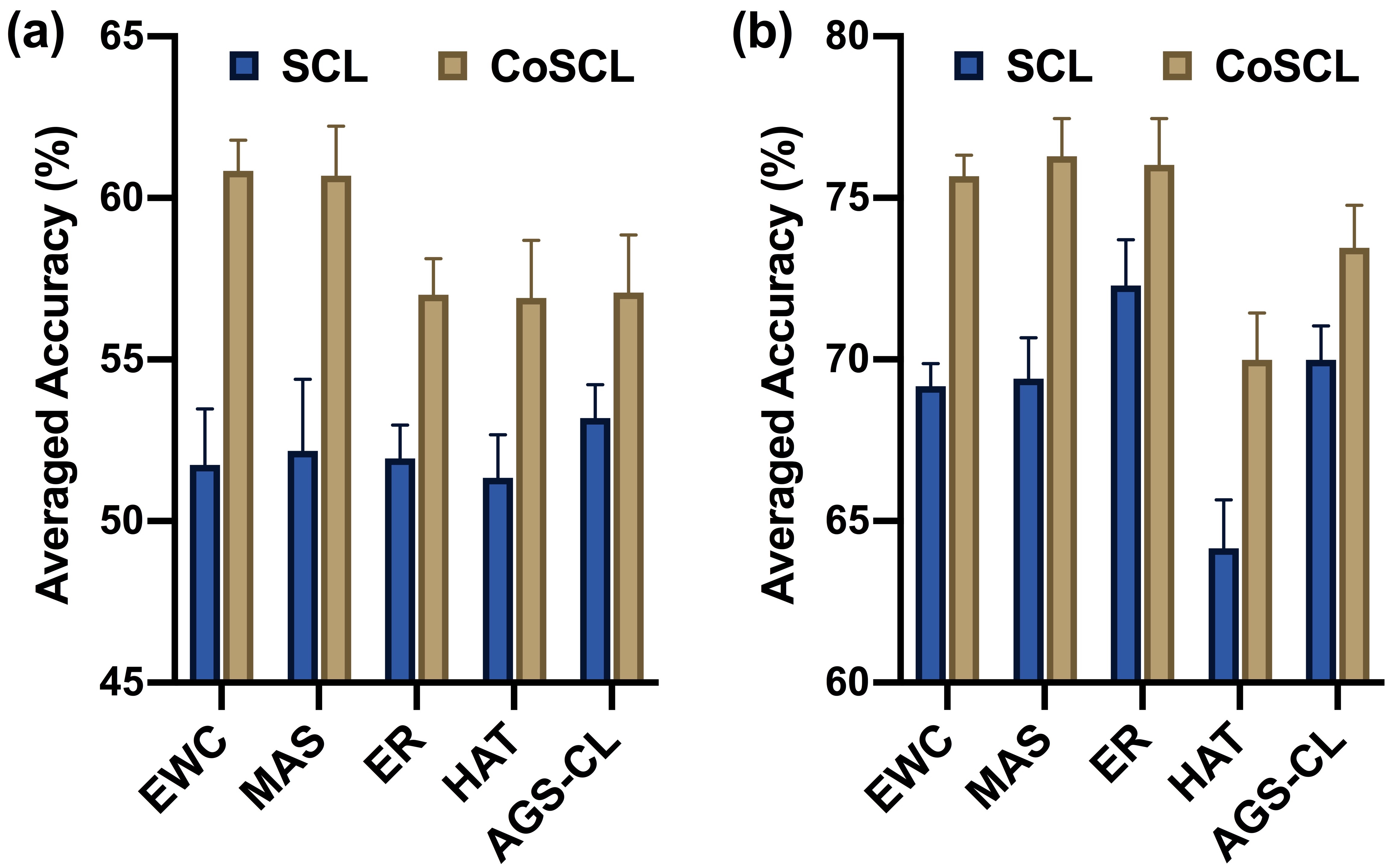} 
    \vspace{-.8cm}
	\caption{Comparison of a single continual learner (SCL) and cooperation of multiple small continual learners with CoSCL (Ours) on (a) CIFAR-100-SC and (b) CIFAR-100-RS.
	} 
	\label{fig:CoSCL_performance}
    \vspace{-.6cm}
\end{wrapfigure}
With a fixed parameter budget, CoSCL can substantially boost a variety of representative continual learning strategies (e.g., up to \textbf{10.64\%} on CIFAR-100-SC, \textbf{9.33\%} on CIFAR-100-RS, \textbf{11.45\%} on CUB-200-2011 and \textbf{6.72\%} on Tiny-ImageNet, detailed in Fig.~\ref{fig:CoSCL_performance} and Table~\ref{table:visual_classification}). The superior performance comes from reducing the errors in both learning plasticity and memory stability by tightening the upper bound, consistent with our theoretical analysis.


Our contributions include: 
(1) We present a unified form of generalization bounds for learning plasticity and memory stability in continual learning;
(2) 
The generalization bounds suggest that the two aspects are not necessarily in conflict, but can be simultaneously enhanced in a compatible parameter space of a well-designed model architecture;
(3) To achieve this goal, we draw inspirations from the biological strategy and propose to cooperate multiple (small) continual learners;
(4) Extensive experiments validate the efficacy and generality of our proposal, which can be adapted to a variety of representative continual learning approaches and improve their performance by a large margin.

\section{Related Work}
\textbf{Continual Learning} requires effective learning of incremental tasks without severe catastrophic forgetting. Representative strategies include weight regularization \cite{kirkpatrick2017overcoming,zenke2017continual,aljundi2018memory}, memory replay \cite{rebuffi2017icarl,wang2021ordisco,wang2021memory}, parameter isolation \cite{serra2018overcoming,jung2020continual} and dynamic architecture \cite{fernando2017pathnet,yan2021dynamically}. These strategies either learn all tasks with a single model, which have to compromise the performance of each task to obtain a shared solution \cite{ramesh2021model}; or allocate parameter subspace for each task to prevent mutual interference, yet limited by scalability.
Several recent work tried to improve continual learning in terms of architecture, such as by using neural architecture search \cite{qin2021bns} or learning an additional set of shared parameters \cite{hurtado2021optimizing}, but to a limited extent. \cite{ramesh2021model} proposed a model zoo that incrementally adds sub-networks to learn new tasks, which had to store a large amount of old training samples.



\textbf{Flatness of Loss Landscape} provides a conceptual explanation of generalization for deep neural networks, which is recently introduced to understand catastrophic forgetting in continual learning \cite{mirzadeh2020understanding,shi2021overcoming,deng2021flattening,cha2020cpr,madaan2021rethinking}. The core idea is that convergence to a smooth region will be more robust to (mild) parameter changes. \cite{mirzadeh2020understanding,dinh2017sharp} analyzed that the forgetting of old tasks in continual learning can be bounded by the variation of parameters between tasks and the eigenvalues of the Hessian matrix, where the lower eigenvalues indicate a flatter curvature of the solution. \cite{cha2020cpr,shi2021overcoming,deng2021flattening} explicitly encouraged the network to find a flat minima and empirically validated its efficacy in continual learning.

\textbf{Ensemble Model} is a powerful architecture to improve generalization, but is still under explored in continual learning. Most current applications focus on learning each single task with a sub-network \cite{rusu2016progressive,aljundi2017expert,wortsman2020supermasks}, which can be seen as a special case of dynamic architecture. The main limitation is that the total amount of parameters (resp., the storage and computational cost) might grow linearly with the number of incremental tasks. \cite{wen2020batchensemble} proposed an efficient ensemble strategy to reduce extra parameter cost for task-specific sub-networks. Similar to ours, a \emph{concurrent} work \cite{doan2022efficient} also observed that ensemble of multiple continual learning models brings huge benefits. They further exploited recent advances of mode connectivity \cite{mirzadeh2020linear} and neural network subspace \cite{wortsman2021learning} to save computational cost, but had to use old training samples \cite{doan2022efficient}. Besides, \cite{liu2020more} achieved more effective weight regularization by ensemble of multiple auxiliary classifiers learned from extra out-of-distribution data (e.g., SVHN \cite{netzer2011reading} for CIFAR-100 \cite{krizhevsky2009learning}).

\textbf{Main Advantages of Our Work} are summarized in three aspects: (1) The generalization bounds presented in our work demonstrate the \emph{direct} link between continual learning performance and flatness of loss landscape (as well as other components). (2) We use a \emph{fixed} number of sub-networks, which are all continual learners rather than single-task learners, and adjust their width accordingly, so no additional or growing parameters are needed. (3) We mainly focus on a restrict setting where old training samples or extra data sources are \emph{not} needed, which is more general and realistic for continual learning.

\section{Preliminary Analysis}
In this section, we first introduce the problem formulation and representative continual learning strategies, and then present the generalization bounds.

\subsection{Problem Formulation}
Let's consider a general setting of continual learning: A neural network with parameter \(\theta\) incrementally learns \(T\) tasks, called a \emph{continual learner}. 
The training set and test set of each task follow the same distribution $\mathbb{D}_t $ ($t=1,2,...,T$), where the training set $D_{t} = \{(x_{t,n}, y_{t,n})\}_{n=1}^{{N}_{t}} $ includes $N_t$ data-label pairs. For classification task, it might include one or several classes. After learning each task, the performance of all the tasks ever seen is evaluated on their test sets.
Although $D_t$ is only available when learning task $t$, an ideal continual learner should behave as if training them jointly. 
To achieve this goal, it is critical to balance learning plasticity of new tasks and memory stability of old tasks. Accordingly, the loss function for continual learning can typically be defined as
\begin{equation}
L_{\rm{CL}}(\theta) = L_{t}(\theta) + \lambda \hat{L}_{1:t-1}(\theta),
\label{eqn.continual_learner}
\end{equation}
where $L_{t}(\cdot)$ is the task-specific loss for learning task $t$ (e.g., cross-entropy for supervised classification), and  $\hat{L}_{1:t-1}(\cdot)$ provides the constraint to achieve a proper trade-off between new and old tasks. 
For example, $\hat{L}_{1:t-1}(\theta) = \sum_{i} I_{1:t-1, i} (\theta_i - \theta_{1:t-1, i}^*)^2$ for weight regularization \cite{kirkpatrick2017overcoming,aljundi2018memory,zenke2017continual}, where $\theta^*_{1:t-1}$ denotes the continually-learned solution for old tasks and $I_{1:t-1}$ indicates the ``importance'' of each parameter. $\hat{L}_{1:t-1}(\theta) = \sum_{k=1}^{t-1}L_{k}(\theta;\hat D_{k})$ for memory replay \cite{rebuffi2017icarl,wang2021ordisco,wang2021memory}, where $\hat D_{k}$ is an approximation of $D_k$ through storing old training samples or learning a generative model. For parameter isolation \cite{serra2018overcoming,jung2020continual}, $\theta = \{ \bigcup_{k=1}^{t-1} \hat{\theta}_k, \hat{\theta}_{\rm{free}} \}$ is dynamically isolated as multiple task-specific subspaces $\hat{\theta}_k$, while $\hat{\theta}_{\rm{free}}$ denotes the ``free'' parameters for current and future tasks. So $\hat{L}_{1:t-1}(\theta)$ usually serves as a sparsity regularizer to save $\hat{\theta}_{\rm{free}}$.
For dynamic architecture \cite{fernando2017pathnet,yan2021dynamically}, $\theta = \{ \bigcup_{k=1}^{t-1} \hat{\theta}_k, \hat{\theta}_t \}$ attempts to add a new subspace $\hat{\theta}_t$ on the basis of the previous ones, and $\hat{L}_{1:t-1}(\theta)$ should limit the amount of extra parameters.

\subsection{Generalization Bound for Continual Learning}


Formally, the goal of continual learning is to find a solution \(\theta\) in a parameter space $\Theta$ that can generalize well over a set of distribution $\mathbb{D}_t$ and $\mathbb{D}_{1:t-1}:=\{\mathbb{D}_k\}_{k=1}^{t-1}$.
Let's consider a bounded loss function $\ell: \mathcal{Y} \times \mathcal{Y} \to [0, c]$ (where $\mathcal{Y}$ denotes a label space and $c$ is the upper bound), such that $\ell(y_1, y_2)=0$ holds if and only if $y_1=y_2$. Then, we can define a population loss over the distribution $\mathbb{D}_t$ by
$\mathcal{E}_{\mathbb{D}_t} (\theta) = \mathbb{E}_{(x, y) \sim \mathbb{D}_t}[\ell(f_{\theta}(x), y)],$
where $f_{\theta}(\cdot)$ is the prediction of an input parameterized by $\theta$.
Likewise, the population loss over the distribution of old tasks is defined by
$\mathcal{E}_{\mathbb{D}_{1:t-1}} (\theta) = \frac{1}{t-1} \sum_{k=1}^{t-1}  \mathbb{E}_{(x, y) \sim \mathbb{D}_k}[\ell(f_{\theta}(x), y)].$
To minimize both $\mathcal{E}_{\mathbb{D}_t} (\theta)$ and $\mathcal{E}_{\mathbb{D}_{1:t-1}} (\theta)$, a continual learning model (i.e., a continual learner) needs to minimize an empirical risk over the current training set $D_{t}$ in a constrained parameter space, i.e., $ \min_{\theta \in \Theta} \hat{\mathcal{E}}_{D_t} (\theta)$. Specifically,
$\hat{\mathcal{E}}_{D_t} (\theta) = \frac{1}{N_t} \sum_{n=1}^{N_t} \ell (f_{\theta}(x_{t,n}), y_{t,n})$,
and the constrained parameter space $\Theta$ depends on the previous experience carried by parameters, data, and/or task labels, so as to prevent catastrophic forgetting.
Likewise, $\hat{\mathcal{E}}_{D_{1:t-1}} (\theta)$ denotes an empirical risk over the old tasks.
In practice, sequential learning of each task by minimizing the empirical risk $\hat{\mathcal{E}}_{D_t} (\theta)$ in $\Theta$ can find multiple solutions, but provides significantly different generalizability on $\mathcal{E}_{\mathbb{D}_t} (\theta)$ and $\mathcal{E}_{\mathbb{D}_{1:t-1}} (\theta)$. Several recent studies suggested that a flatter solution is more robust to catastrophic forgetting \cite{mirzadeh2020understanding,shi2021overcoming,deng2021flattening,cha2020cpr,madaan2021rethinking}. To find such a flat solution, we define a robust empirical risk by the worst case of the neighborhood in parameter space as $\hat{\mathcal{E}}_{D_t}^b (\theta) := \rm{max}_{\lVert \Delta \rVert \leq b} \hat{\mathcal{E}}_{D_t} (\theta +\Delta)$ \cite{cha2021swad}, where $b$ is the radius around $\theta$ and $\lVert \cdot \rVert$ denotes the L2 norm, likewise for the old tasks as
$\hat{\mathcal{E}}_{D_{1:t-1}}^b (\theta) := \rm{max}_{\lVert \Delta \rVert \leq b} \hat{\mathcal{E}}_{D_{1:t-1}} (\theta +\Delta)$. Then, solving the constrained robust empirical risk minimization, i.e., $ \min_{\theta \in \Theta} \hat{\mathcal{E}}_{D_t}^b (\theta)$, will find a near solution of a flat optimum showing better generalizability. In particular, the minima found by the empirical loss $\hat{\mathcal{E}}_{D_t} (\theta) $ will also be the minima of $\hat{\mathcal{E}}_{D_t}^b (\theta)$ if the ``radius'' of its loss landscape is sufficiently wider than $b$. 
Intuitively, such a flat solution helps to mitigate catastrophic forgetting since it is more robust to parameter changes. 

However, this connection is not sufficient. If a new task is too different from the old tasks, the parameter changes to learn it well might be much larger than the ``radius'' of the old minima, resulting in catastrophic forgetting. On the other hand, staying around the old minima is not a good solution for the new task, limiting learning plasticity. Let $\mathcal{E}_{\mathbb{D}_t} (\theta_{1:t})$ and $\mathcal{E}_{\mathbb{D}_{1:t-1}} (\theta_{1:t})$ denote the generalization errors of performing the new task and old tasks, respectively.
Inspired by the PAC-Bayes theory \cite{mcallester1999pac} and previous work in domain generalization \cite{cha2021swad,ben2010theory}, we first present the upper bounds of these two errors as follows (please see a complete proof in Appendix A):

\begin{proposition} \label{errorbound}
Let $\Theta$ be a cover of a parameter space with VC dimension $d$. 
If $\mathbb{D}_1,\cdots,\mathbb{D}_t$ are the distributions of the continually learned $1:t$ tasks,
then for any $\delta \in (0,1)$ with probability at least $1-\delta$,
for every solution $\theta_{1:t}$ of the continually learned $1:t$ tasks in parameter space $\Theta$, i.e., $\theta_{1:t} \in \Theta$:
\begin{small}
\begin{align} \label{pro1-1}
 \mathcal{E}_{\mathbb{D}_t} (\theta_{1:t}) <  \hat{\mathcal{E}}_{D_{1:t-1}}^b (\theta_{1:t}) + \frac{1}{2(t-1)}\sum_{k=1}^{t-1} \rm{Div}(\mathbb{D}_\textit{k}, \mathbb{D}_\textit{t}) + \sqrt{\frac{{\textit d}\ln (\textit N_{1:\textit t-1}/\textit d) + \ln (1/\delta)}{\textit N_{1:\textit t-1}} } ,
\end{align}
\vspace{-.2cm}
\begin{align} \label{pro1-2}
 \mathcal{E}_{\mathbb{D}_{1:t-1}} (\theta_{1:t})  <  \hat{\mathcal{E}}_{D_{t}}^b (\theta_{1:t}) + \frac{1}{2(t-1)}\sum_{k=1}^{t-1} \rm{Div}(\mathbb{D}_\textit{t}, \mathbb{D}_\textit{k}) + \sqrt{\frac{\textit{d}\ln (\textit{N}_{\textit{t}}/\textit{d}) + \ln (1/\delta)}{\textit{N}_{\textit{t}}} },
\end{align}
\end{small}
\noindent where $\rm{Div}(\mathbb{D}_\textit{i}, \mathbb{D}_\textit{j}) := 2 \sup_{\textit{h} \in \mathcal{H}} |\mathcal{P}_{ \mathbb{D}_\textit{i}}(\textit{I}(\textit{h})) - \mathcal{P}_{ \mathbb{D}_\textit{j}}(\textit{I}(\textit{h}))|$ is the $\mathcal{H}$-divergence for the distribution $\mathbb{D}_i$ and $\mathbb{D}_j$ ($I(h)$ is the characteristic function). $N_{1:t-1}=\sum_{k=1}^{t-1} N_k$ is the total number of training samples over all old tasks.
\end{proposition}



\begin{figure}[t]
	\centering
	\includegraphics[width=1\columnwidth]{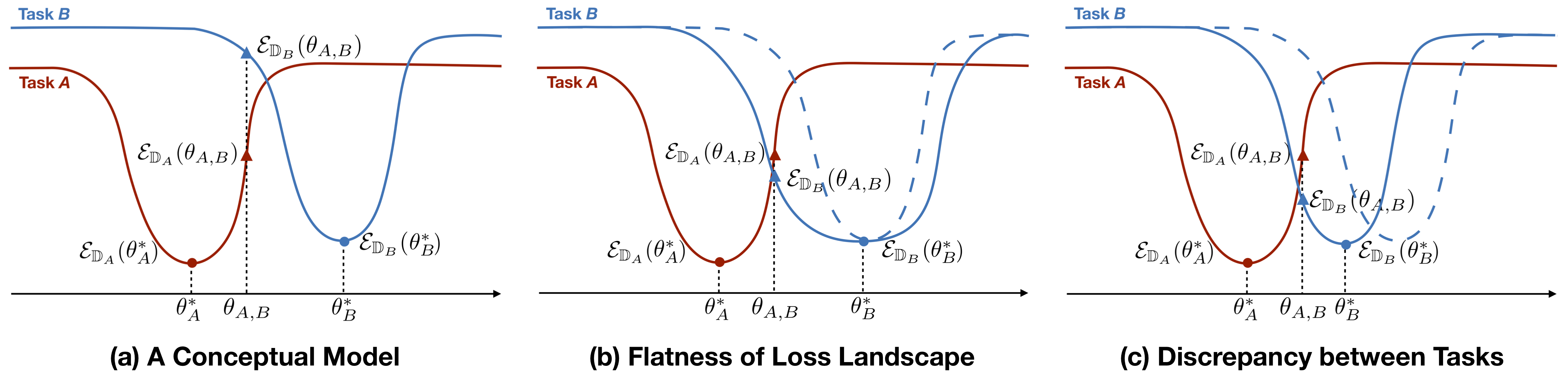} 
    \vspace{-.4cm}
	\caption{A conceptual model of two tasks for our theoretical analysis (the learning order of Task $A$ and Task $B$ does not matter). The dashed line in (b) and (c) is the original solution in (a), where finding a flatter solution or reducing the discrepancy between tasks help to mitigate the generalization errors of a shared solution $\theta_{A,B}$.}
	\label{fig:theory_demo}
     \vspace{-.1cm}
\end{figure}

\begin{figure}[th]
	\centering
	\includegraphics[width=0.80\columnwidth]{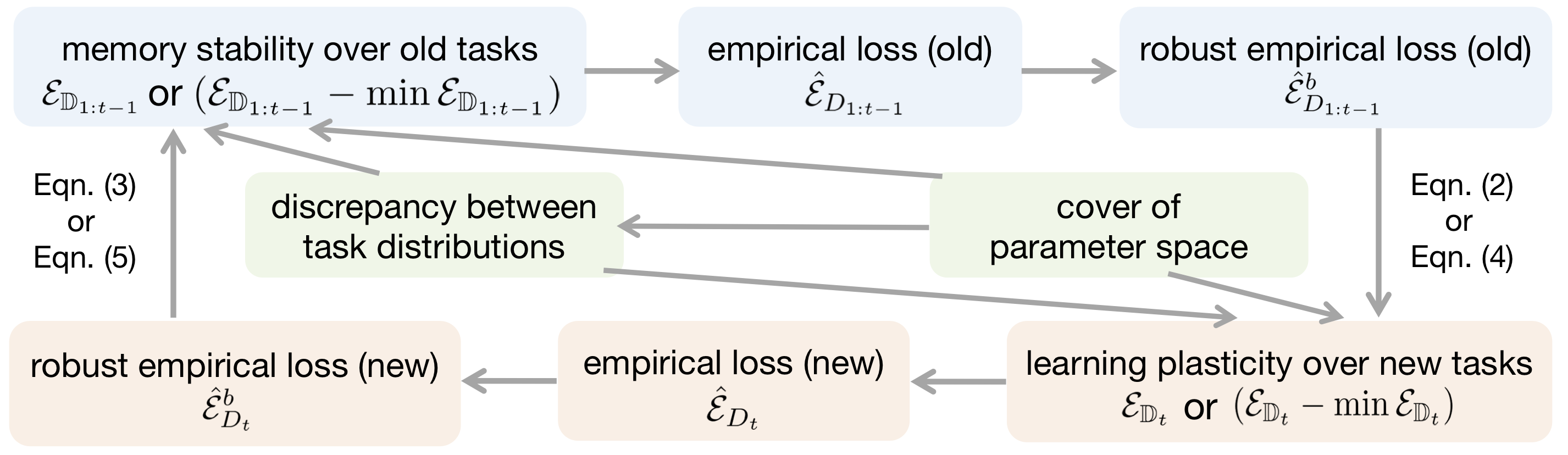} 
    \vspace{-.1cm}
	\caption
	{Illustration of simultaneously promoting learning plasticity and memory stability in continual learning, where arrows represent the tightening process.} 
	\label{fig:error}
    \vspace{-.4cm}
\end{figure}


It can be concluded from Proposition~\ref{errorbound} that, the generalization errors over the new task and old tasks are uniformly constrained by three components: (1) \emph{discrepancy between task distributions}; (2) \emph{flatness of loss landscape}; and (3) \emph{cover of parameter space}.
By the optimal solution for (robust) empirical loss, we further demonstrate that the generalization gaps of the new task and old tasks are upper bounded as follows (the proof is detailed in Appendix A):



\begin{proposition}\label{gapbound}
Let $\hat \theta_{1:t}^{b}$ denotes the optimal solution of the continually learned $1:t$ tasks by
robust empirical risk minimization over the current task, i.e., $\hat \theta_{1:t}^{b} = \arg \min_{\theta \in \Theta} \hat{\mathcal{E}}_{D_t}^b (\theta)$, where $\Theta$ denotes a cover of a parameter space with VC dimension $d$. Then for any $\delta \in (0,1)$, with probability at least $1-\delta$:
\begin{small}
\begin{align} \label{pro2-1}
\begin{split}
 \mathcal{E}_{\mathbb{D}_{t}} (\hat \theta_{1:t}^{b}) - \min_{\theta \in \Theta} \mathcal{E}_{\mathbb{D}_{t}} (\theta ) 
\leq \min_{\theta \in \Theta} \hat{\mathcal{E}}_{D_{1:t-1}}^b (\theta) -  \min_{\theta \in \Theta} \hat{\mathcal{E}}_{D_{1:t-1}} (\theta) + \frac{1}{t-1}\sum_{k=1}^{t-1} \rm{Div}( \mathbb{D}_\textit{k}, \mathbb{D}_\textit{t}) + \lambda_{1},
 \end{split}
\end{align}
\vspace{-.6cm}
\begin{align}\label{pro2-2}
\begin{split}
 \mathcal{E}_{\mathbb{D}_{1:t-1}} (\hat \theta_{1:t}^{b}) - \min_{\theta \in \Theta} \mathcal{E}_{\mathbb{D}_{1:t-1}} (\theta ) 
\leq \min_{\theta \in \Theta} \hat{\mathcal{E}}_{D_t}^b (\theta) -  \min_{\theta \in \Theta} \hat{\mathcal{E}}_{D_t} (\theta) + \frac{1}{t-1}\sum_{k=1}^{t-1} \rm{Div}(\mathbb{D}_\textit{t}, \mathbb{D}_\textit{k}) + \lambda_{2},
 \end{split}
\end{align}
\end{small}
where $\lambda_{1} = 2\sqrt{\frac{d\ln (N_{1:t-1}/d) + \ln (2/\delta)}{N_{1:t-1}} } $, $\lambda_{2} = 2\sqrt{\frac{d\ln (N_{t}/d) + \ln (2/\delta)}{N_{t}} } $, and $\rm{Div}(\mathbb{D}_\textit{i}, \mathbb{D}_\textit{j}) := 2 \sup_{\textit{h} \in \mathcal{H}} |\mathcal{P}_{ \mathbb{D}_\textit{i}}(\textit{I}(\textit{h})) - \mathcal{P}_{ \mathbb{D}_\textit{j}}(\textit{I}(\textit{h}))|$ is the $\mathcal{H}$-divergence for the distribution $\mathbb{D}_i$ and $\mathbb{D}_j$ ($I(h)$ is the characteristic function).
\end{proposition}

Likewise, the generalization gaps over the new and old tasks are also constrained by the three components above.
In particular, learning plasticity and memory stability in continual learning can be simultaneously promoted by using a more compatible parameter space, as illustrated in Fig.~\ref{fig:error}. Specifically, compatibility with the new task can facilitate a smaller robust empirical risk on the old tasks as well as improve task discrepancy, then tightening the generalization bound for learning plasticity through Eqn.~(\ref{pro1-1})/Eqn.~(\ref{pro2-1}), and vice versa tightening the generalization bound for memory stability through Eqn.~(\ref{pro1-2})/Eqn.~(\ref{pro2-2}).

\section{Method}

\begin{figure}[t]
	\centering
	\includegraphics[width=1\columnwidth]{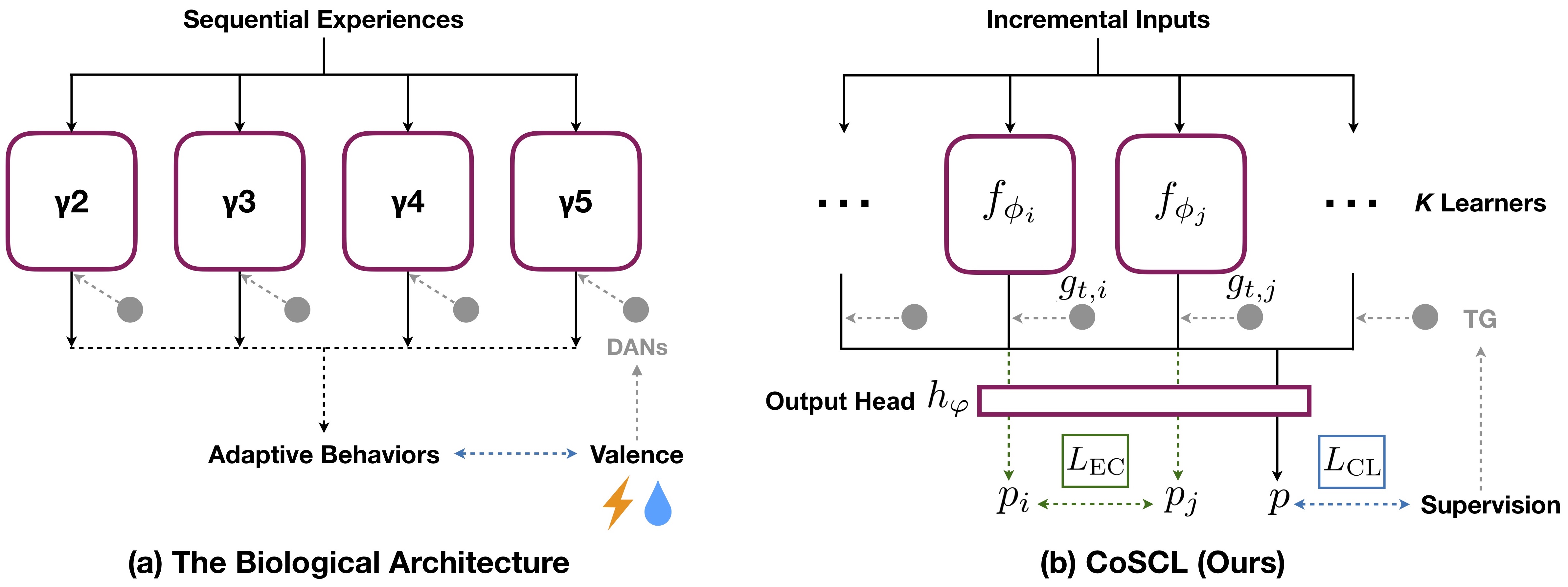} 
    \vspace{-.4cm}
	\caption{(a) Fruit flies learn sequential experiences with multiple parallel compartments ($\gamma$ 2-5), under the modulation of dopaminergic neurons (DANs) that convey valence. The figure is modified from \cite{cohn2015coordinated}.
	(b) Inspired by the biological learning system, we propose a general strategy of cooperating multiple (small) continual learners.}
	\label{fig:bio_CoSCL_model}
  \vspace{-.3cm}
\end{figure}

Unlike artificial neural networks, the robust biological learning system, such as that of fruit flies, processes sequential experiences with multiple parallel compartments (i.e, sub-networks) \cite{aso2014neuronal,cohn2015coordinated}. These compartments are modulated by dopaminergic neurons (DANs) that convey valence (i.e., supervised signals), and their outputs are integrated in a weighted-sum fashion to guide adaptive behaviors \cite{aso2014neuronal,cohn2015coordinated,modi2020drosophila} (detailed in Fig.~\ref{fig:bio_CoSCL_model}, a). 
Inspired by this, we propose to cooperate multiple (small) continual learners as a simple yet effective method for continual learning. We present our proposal in Sec.~\ref{sec4.1}, and validate this idea both theoretically (Sec. 4.2) and empirically (Sec. 5). 


\subsection{Cooperation of (Small) Continual Learners}\label{sec4.1}
Instead of learning all tasks with a single continual learner, we design a bio-inspired architecture to coordinate multiple continual learners. Specifically, each continual learner is implemented with a sub-network $f_{\phi_i}( \cdot ), i=1,...,K$ in a parameter space for learning all incremental tasks, where the dedicated output head is removed and the output of the previous layer is weighted by a set of learnable parameters (usually a fully connected layer). Then, these outputs are fed into a shared output head $h_{\varphi}(\cdot)$ for prediction. For a regular classifier, this is equivalent to making predictions on a weighted-sum of feature representations, so we refer to this strategy as \emph{feature ensemble} (FE).

When task labels are available, our architecture can more effectively incorporate task-specific information by learning an additional set of \emph{task-adaptive gates} (TG) for each continual learner's output. Such a gate is defined as $g_{t, i} = \sigma (s \cdot \alpha_{t, i})$ for learner $i$ to perform task $t$, where $\alpha_{t, i}$ is a learnable parameter, $s$ is a scale factor and $\sigma$ denotes the sigmoid function. Therefore, the final prediction becomes $p(\cdot) =h_{\varphi}(\sum_{i=1}^K g_{t,i} f_{\phi_i}( \cdot))$, and all optimizable parameters include $\bar{\theta} = \{ \bigcup_{i=1}^{K}\phi_i, \bigcup_{i=1}^{K}\alpha_{t,i}, \varphi\}$. 

To strengthen the advantage of feature ensemble, we encourage to cooperate the continual learners by penalizing differences in the predictions of their feature representations (e.g., $\textbf{\textit{p}}_i$ and $\textbf{\textit{p}}_j$). We choose the widely-used Kullback Leibler (KL) divergence and define an \emph{ensemble cooperation} (EC) loss as
\begin{equation}
\begin{split}
L_{\rm{EC}}(\bar{\theta}) & =  \frac{1}{K} \sum_{i = 1, j \neq i}^{K} D_{KL}(\textbf{\textit{p}}_i|| \textbf{\textit{p}}_j) 
=\frac{1}{K} \frac{1}{N_t} \sum_{i = 1, j \neq i}^{K} \sum_{n = 1}^{N_t} p_i(x_{t,n}) \log \frac{p_i(x_{t,n})}{p_j(x_{t,n})}\\
& = \frac{1}{K} \frac{1}{N_t} \sum_{i = 1, j \neq i}^{K} \sum_{n = 1}^{N_t} h_{\varphi}( g_{t,i} f_{\phi_i}(x_{t,n})) \log \frac{h_{\varphi}( g_{t,i} f_{\phi_i}(x_{t,n}))}{h_{\varphi}( g_{t,j} f_{\phi_j}(x_{t,n}))}.
\end{split}
\label{eqn.ec_loss}
\end{equation}
In practice, we reduce the sub-network width to save parameters, so we call our method ``Cooperation of Small Continual Learners (CoSCL)''. Taking Eqn.~(\ref{eqn.continual_learner}) and Eqn.~(\ref{eqn.ec_loss}) together, the objective of CoSCL is defined as
\begin{equation}
L_{\rm{CoSCL}}(\bar{\theta}) = \textit{L}_{\rm{CL}}(\bar{\theta} ) + \gamma \textit{L}_{\rm{EC}}( \bar{\theta} ).
\label{eqn.CoSCL}
\end{equation}

\subsection{Theoretical Explanation}
Here we provide a theoretical explanation of how cooperating multiple continual learners can mitigate the generalization gaps in continual learning:
\begin{proposition} \label{coscl}
Let $\{\Theta_i \in  \mathbb{R}^{r}\}_{i=1}^{K}$ be a set of $K$ parameter spaces ($K>1$ in general), 
$d_i$ be a VC dimension of $\Theta_i$, and $\Theta = \cup_{i=1}^{K}\Theta_i$ with VC dimension $d$.
Based on Proposition~\ref{gapbound}, for $\hat \theta_{1:t}^{b} = \arg \min_{\bar{\theta} \in \Theta} \hat{\mathcal{E}}_{D_t}^b (\bar{\theta})$, the upper bound of generalization gap is further tighter with
\begin{small}
\begin{align}
    \lambda_{1} = \max_{i \in [1,K]} \sqrt{\frac{d_i\ln (N_{1:t-1}/d_i) + \ln (2K/\delta)}{N_{1:t-1}} }+ \sqrt{\frac{d\ln (N_{1:t-1}/d) + \ln (2/\delta)}{N_{1:t-1}} },
\end{align}
\vspace{-.2cm}
\begin{align}
    \lambda_{2} = \max_{i \in [1,K]} \sqrt{\frac{d_i\ln (N_{t}/d_i) + \ln (2K/\delta)}{N_{t}} } + \sqrt{\frac{d\ln (N_{t}/d) + \ln (2/\delta)}{N_{t}} }.
\end{align}
\end{small}
\end{proposition}
Comparing Proposition~\ref{coscl} and Proposition~\ref{gapbound}, we conclude that cooperating $K$ continual learners facilitates a smaller generalization gap over the new and old tasks in continual learning than a single one. 
Due to the space limit, we leave more details of Proposition~\ref{coscl} in Appendix A, where we also analyze how a compatible parameter space of a well-designed model architecture can improve the discrepancy between task distributions, thus further tightening the generalization bounds. Next, we empirically validate our proposal as detailed below.


\section{Experiment}
In this section, we extensively evaluate CoSCL on visual classification tasks. All results are averaged over 5 runs with different random seeds and task orders.

\textbf{Benchmark:} We consider four representative continual learning benchmarks. The first two are with CIFAR-100 dataset \cite{krizhevsky2009learning}, which includes 100-class colored images of the size \(32 \times 32\). All classes are split into 20 incremental tasks, based on random sequence (RS) or superclass (SC).
The other two are with larger-scale datasets, randomly split into 10 incremental tasks: CUB-200-2011 \cite{wah2011caltech} includes 200 classes and 11,788 bird images of the size $224 \times 224$, and is split as 30 images per class for training while the rest for testing. Tiny-ImageNet \cite{delange2021continual} is derived from iILSVRC-2012 \cite{russakovsky2015imagenet}, consisting of 200-class natural images of the size \(64 \times 64\). 

\begin{wrapfigure}{r}{0.48\textwidth}
    \vspace{-.6cm}
	\centering
	\includegraphics[width=0.48\columnwidth]{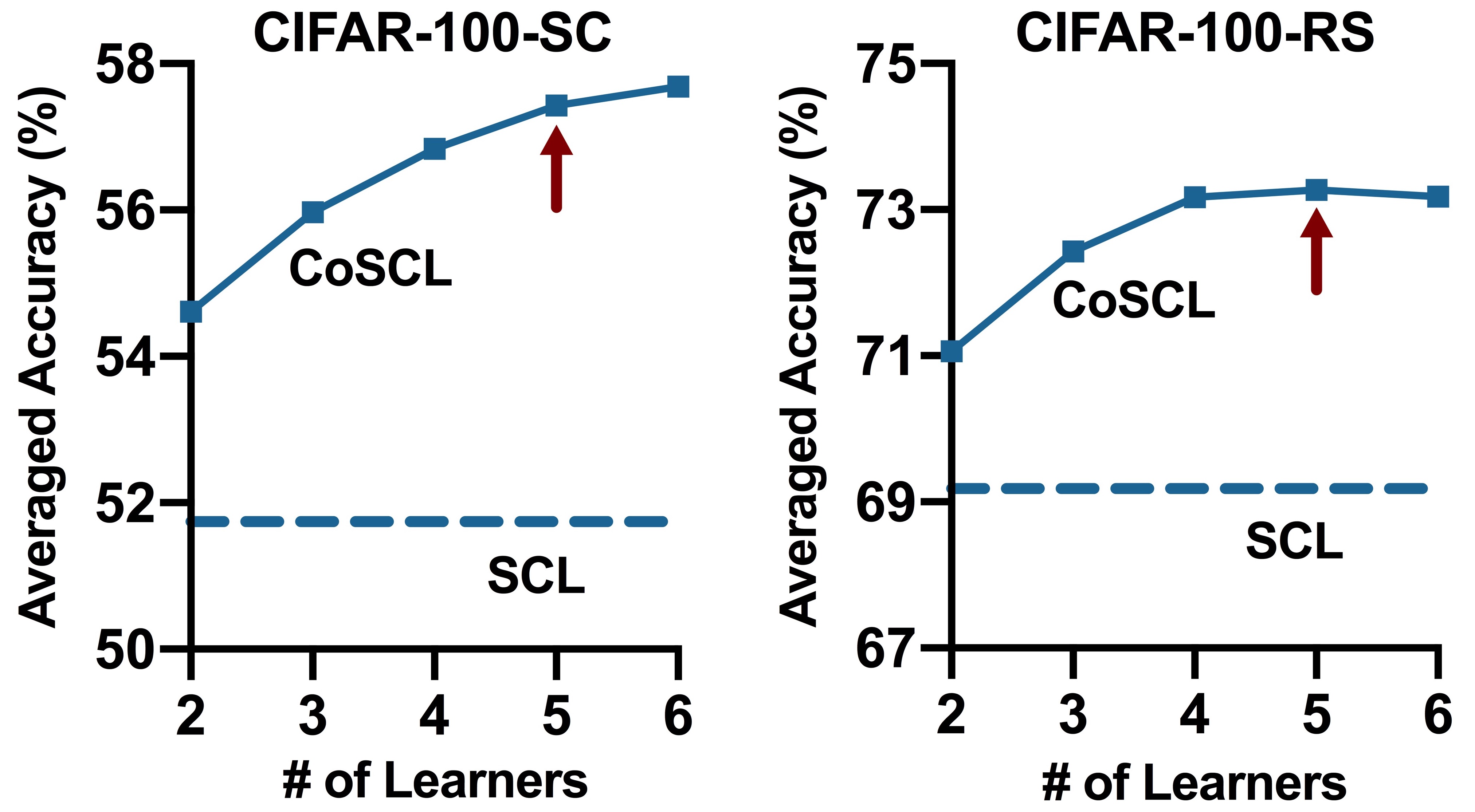} 
    \vspace{-.7cm}
	\caption{Trade-off between learner number and width with a fixed parameter budget. }
	\label{fig:numbe_width}
    \vspace{-.7cm}
\end{wrapfigure}

\textbf{Implementation:} We mainly focus on the task-incremental setting used in \cite{wang2021afec,jung2020continual,cha2020cpr,serra2018overcoming} and follow their implementation for most experiments if not specified. For all the baselines, we apply a 6-layer CNN architecture for CIFAR-100-SC and CIFAR-100-RS, and an AlexNet-based architecture for CUB-200-2011 and Tiny-ImageNet.\footnote{A concurrent work observed that the regular CNN architecture indeed achieves better continual learning performance than more advanced architectures such as ResNet and ViT with the same amount of parameters \cite{mirzadeh2022architecture}.} 
Since our method consists of multiple continual learners, we use a similar architecture for each sub-network and accordingly reduce the width (i.e., using fewer channels) to keep the total number of parameters comparable to other baselines, so as to make the comparison as fair as possible.
Then, there is an intuitive trade-off between the number and width of learners.
According to our theoretical analysis in Proposition~\ref{errorbound} and \ref{gapbound}, the choice for the number of learners (i.e., parameter spaces) $K$ is \emph{independent} of the training data distribution under a limited parameter budget. Also, we empirically validate that this trade-off is only moderately sensitive (see Fig.~\ref{fig:numbe_width}). So we simply set $K=5$ for all experiments. 
The learners' training differs only in random initialization of the parameters. The implementations are further detailed in Appendix B.


\renewcommand\arraystretch{1.5}
\begin{table*}[t]
	\centering
	\caption{
Averaged accuracy (\%) of all the tasks learned so far in continual learning ($A_t$ for $t$ tasks). All results are cited from \cite{wang2021afec,jung2020continual,cha2020cpr} or reproduced from their officially-released code for a fair comparison. CoSCL cooperates 5 continual learners with similar architectures as other baselines, while reducing the sub-network width accordingly to keep the total amount of parameters comparable.}
    \vspace{-.2cm}
	\smallskip
	\resizebox{0.95\textwidth}{!}{ 
	\begin{tabular}{lcccccccc}
		\specialrule{0.01em}{1.2pt}{1.5pt}
		 \multicolumn{1}{c}{} & \multicolumn{2}{c}{CIFAR-100-SC} & \multicolumn{2}{c}{CIFAR-100-RS} & \multicolumn{2}{c}{CUB-200-2011} & \multicolumn{2}{c}{Tiny-ImageNet}\\
       Methods & \(A_{10}\) & \(A_{20}\) & \(A_{10}\) & \(A_{20}\) & \(A_{5}\) & \(A_{10}\) & \(A_{5}\) & \(A_{10}\)  \\
       \specialrule{0.01em}{1.2pt}{1.7pt}
        SI \cite{zenke2017continual}&52.20 \tiny{\(\pm 4.37\)}&51.97 \tiny{\(\pm 2.07\)}&68.72 \tiny{\(\pm 1.11\)}&69.21 \tiny{\(\pm 0.77\)}&33.08 \tiny{\(\pm 4.05\)}&42.03 \tiny{\(\pm 3.06\)}&45.61 \tiny{\(\pm 2.05\)}&46.00 \tiny{\(\pm 1.13\)}\\
        RWALK \cite{chaudhry2018riemannian}&50.51 \tiny{\(\pm 4.53\)}&49.62 \tiny{\(\pm 3.28\)}&66.02 \tiny{\(\pm 1.89\)}&66.90 \tiny{\(\pm 0.29\)}&32.56 \tiny{\(\pm 3.76\)}&41.94 \tiny{\(\pm 2.35\)}&49.69 \tiny{\(\pm 1.47\)}&48.12 \tiny{\(\pm 0.96\)} \\
       P\&C \cite{schwarz2018progress}&53.48  \tiny{\(\pm 2.79\)}&52.88 \tiny{\(\pm 1.68\)}&70.10 \tiny{\(\pm 1.22\)}&70.21 \tiny{\(\pm \)1.22}&33.88 \tiny{\(\pm 4.48\)}&42.79 \tiny{\(\pm 3.29\)}&51.71 \tiny{\(\pm 1.58\)}&50.33 \tiny{\(\pm 0.86\)}\\
      \specialrule{0.01em}{1.2pt}{1.7pt}
       EWC \cite{kirkpatrick2017overcoming}&52.25 \tiny{\(\pm 2.99 \)}&51.74 \tiny{\(\pm 1.74\)}&68.72 \tiny{\(\pm 0.24\)}&69.18 \tiny{\(\pm 0.69\)}&32.90 \tiny{\(\pm 2.98\)}&42.29 \tiny{\(\pm 2.34\)}&50.92 \tiny{\(\pm 1.86\)}&48.38 \tiny{\(\pm 0.86\)}\\
      \ \emph{w/} AFEC \cite{wang2021afec} & 56.28 \tiny{\(\pm 3.27\)}&55.24 \tiny{\(\pm 1.61\)}& 72.36 \tiny{\(\pm 1.23\)}& 72.29 \tiny{\(\pm 1.07\)}&34.36 \tiny{\(\pm 4.39\)}&43.05 \tiny{\(\pm 3.00\)}&51.34 \tiny{\(\pm 1.62\)}&50.58 \tiny{\(\pm 0.74\)}\\
       \ \emph{w/} CPR \cite{cha2020cpr}&54.60 \tiny{\(\pm 2.51\)}&53.37 \tiny{\(\pm 2.06\)}&71.12 \tiny{\(\pm 1.82\)}&70.25 \tiny{\(\pm 1.33\)}&33.36 \tiny{\(\pm 3.25\)}&42.51 \tiny{\(\pm 2.31\)}&50.12 \tiny{\(\pm 1.43\)}&50.29 \tiny{\(\pm 0.89\)}\\
       \rowcolor{black!20}
       \ \emph{w/} CoSCL (Ours)  &\textbf{62.89} \tiny{\(\pm 3.05\)}&\textbf{60.84} \tiny{\(\pm 0.95\)}&\textbf{78.08} \tiny{\(\pm 1.25\)}&76.05 \tiny{\(\pm 0.65\)}&\textbf{44.35} \tiny{\(\pm 3.59\)}&48.53 \tiny{\(\pm 2.21\)}&\textbf{56.10} \tiny{\(\pm 1.77\)}&55.10 \tiny{\(\pm 1.02\)}\\
     \specialrule{0.01em}{1.2pt}{1.7pt}
       MAS \cite{aljundi2018memory} &52.76 \tiny{\(\pm 2.85\)}&52.18 \tiny{\(\pm 2.22\)}&67.60 \tiny{\(\pm 1.85\)}&69.41 \tiny{\(\pm 1.27\)}&31.68 \tiny{\(\pm 2.37\)}&42.56 \tiny{\(\pm 1.84\)}&49.69 \tiny{\(\pm 1.50\)}&50.20 \tiny{\(\pm 0.82\)}\\
       \ \emph{w/} AFEC \cite{wang2021afec} &55.26 \tiny{\(\pm 4.14\)}&54.89 \tiny{\(\pm 2.23\)}&69.57 \tiny{\(\pm 1.73\)}&71.20 \tiny{\(\pm 0.70\)}&34.08 \tiny{\(\pm 3.80\)}&42.93 \tiny{\(\pm 3.51\)}&51.35 \tiny{\(\pm 1.75\)}&50.90 \tiny{\(\pm 1.08\)}\\
       \ \emph{w/} CPR \cite{cha2020cpr}&52.90 \tiny{\(\pm 1.62\)}&53.63 \tiny{\(\pm 1.31\)}&70.69 \tiny{\(\pm 1.85\)}&72.06 \tiny{\(\pm 1.86\)}&33.49 \tiny{\(\pm 2.46\)}&43.07 \tiny{\(\pm 2.56\)}&50.82 \tiny{\(\pm 1.41\)}&51.24 \tiny{\(\pm 1.26\)}\\
        \rowcolor{black!20}
      \ \emph{w/} CoSCL (Ours)  &62.55 \tiny{\(\pm 1.94\)}&60.69 \tiny{\(\pm 1.53\)}&76.93 \tiny{\(\pm 1.94\)}&\textbf{76.29} \tiny{\(\pm 2.33\)}&43.67 \tiny{\(\pm 3.73\)}&\textbf{49.48} \tiny{\(\pm 2.40\)}&55.43 \tiny{\(\pm 1.48\)}&\textbf{55.11} \tiny{\(\pm 0.89\)}\\

      \specialrule{0.01em}{1.2pt}{1.7pt}
	\end{tabular}
	}
	\label{table:visual_classification}
	\vspace{-.1cm}
\end{table*}

\begin{figure}[t]
	\centering
	\includegraphics[width=0.90\columnwidth]{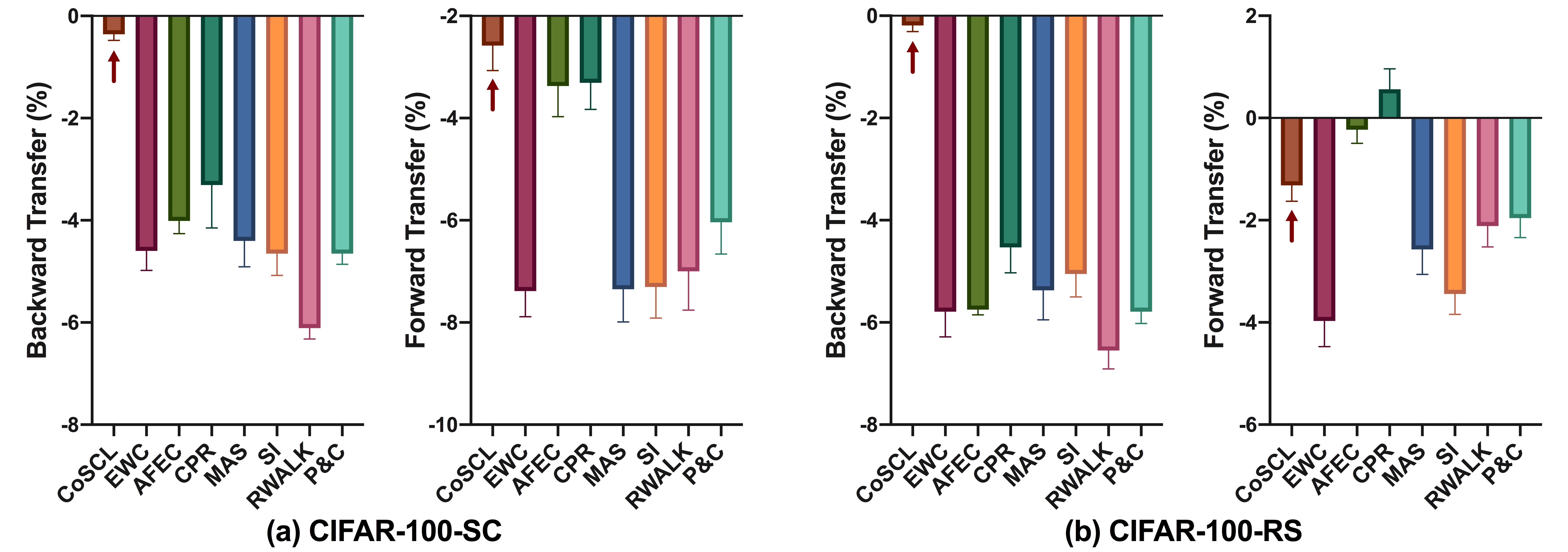} 
    \vspace{-.2cm}
	\caption{Comparison of backward transfer (BWT) and forward transfer (FWT). } 
	\label{fig:knowledge_transfer}
    \vspace{-.5cm}
\end{figure}

\textbf{Overall Performance:} 
We first adapt CoSCL to representative continual learning strategies, including weight regularization such as EWC \cite{kirkpatrick2017overcoming} and MAS \cite{aljundi2018memory}, parameter isolation such as HAT \cite{serra2018overcoming} and AGS-CL \cite{jung2020continual}, and experience replay (ER) of old training samples (20 images per class) \cite{riemer2018learning}. As shown in Fig.~\ref{fig:CoSCL_performance}, our proposal that cooperates multiple continual learners with narrower sub-networks can largely improve their performance. Then, we compare with the state-of-the-art (SOTA) methods under a realistic restriction that old training samples or additional data sources are \emph{not} available, as detailed below.

First, we compare with the SOTA methods that can be plug-and-play with weight regularization baselines, such as AFEC \cite{wang2021afec} and CPR \cite{cha2020cpr}. AFEC \cite{wang2021afec} encouraged the network parameters to resemble the optimal solution for each new task to mitigate potential negative transfer, while CPR \cite{cha2020cpr} added a regularization term that maximized the entropy of output probability to find a flat minima. In contrast, CoSCL can more effectively improve weight regularization baselines by up to 10.64\% on CIFAR-100-SC, 9.33\% on CIFAR-100-RS, 11.45\% on CUB-200-2011 and 6.72\% on Tiny-ImageNet, and achieve the new SOTA performance (detailed in Table \ref{table:visual_classification}). 

\begin{wraptable}{r}{0.48\textwidth}
	\centering
    \vspace{-0.3cm}
	\caption{Averaged accuracy (\%) of architecture-based methods on CIFAR-100-RS. Here we use EWC as the default continual learning method for CoSCL.}
	\smallskip
	\resizebox{0.48\textwidth}{!}{ 
	\begin{tabular}{lccc}
		\specialrule{0.01em}{1.2pt}{1.5pt}
       Methods & \# Param & 20-split & 50-split \\
       \specialrule{0.01em}{1.2pt}{1.7pt}
       HAT \cite{serra2018overcoming} &6.8M &76.96 &80.46 \\
       MARK \cite{hurtado2021optimizing} &4.7M &78.31 &-- \\
       BNS \cite{qin2021bns}&6.7M &-- &82.39 \\
       \rowcolor{black!20}
       CoSCL (Ours) &4.6M &\textbf{79.43} \tiny{\(\pm 1.01\)}&\textbf{87.88} \tiny{\(\pm 1.07\)}\\
      \specialrule{0.01em}{1.2pt}{1.7pt}
	\end{tabular}
	}
	\label{table:architecture_method}
	\vspace{-.9cm}
\end{wraptable}

At the same time, we consider the SOTA methods that improve continual learning in terms of architecture, such as BNS \cite{qin2021bns} and MARK \cite{hurtado2021optimizing}. BNS applied neural structure search to build a network for preventing catastrophic forgetting and promoting knowledge transfer, while MARK achieved this goal by learning an additional set of shared weights among tasks.\footnote{They both are performed against a similar AlexNet-based architecture.} With a similar or smaller parameter budget, ours largely outperforms the two recent strong baselines (see Table~\ref{table:architecture_method}).


\begin{figure}[t]
	\centering
	\includegraphics[width=0.85\columnwidth]{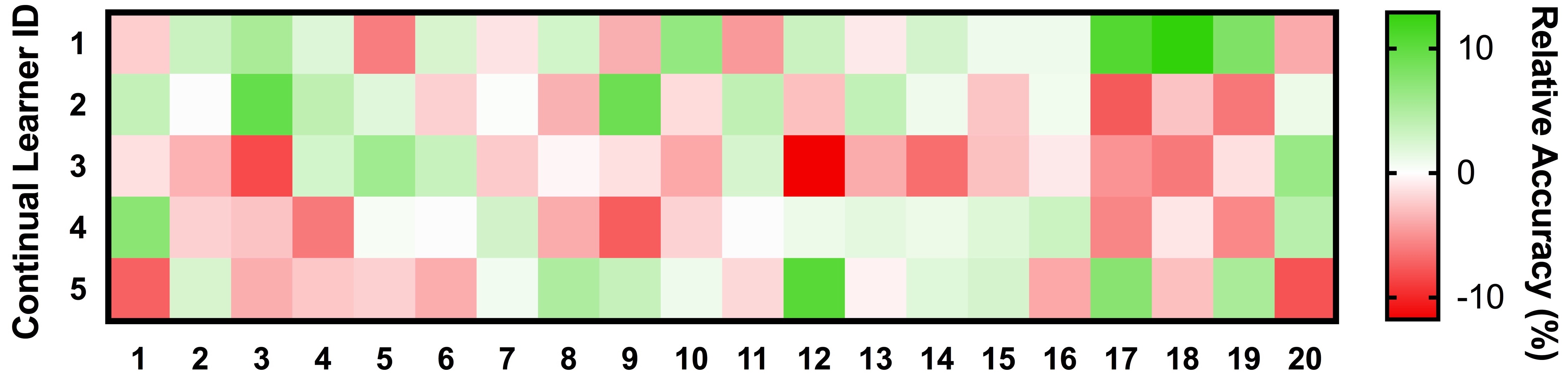}
    \vspace{-.2cm}
	\caption{Diversity of expertise in CIFAR-100-SC. The relative accuracy is calculated by subtracting the performance of each learner from the averaged performance of all learners. An accuracy gap of about $10\% \sim 20\%$ exists between the best and the worst.}
	\label{fig:task_diversity}
    \vspace{-.4cm}
\end{figure}

\begin{figure}[th]
	\centering
	\includegraphics[width=0.78\columnwidth]{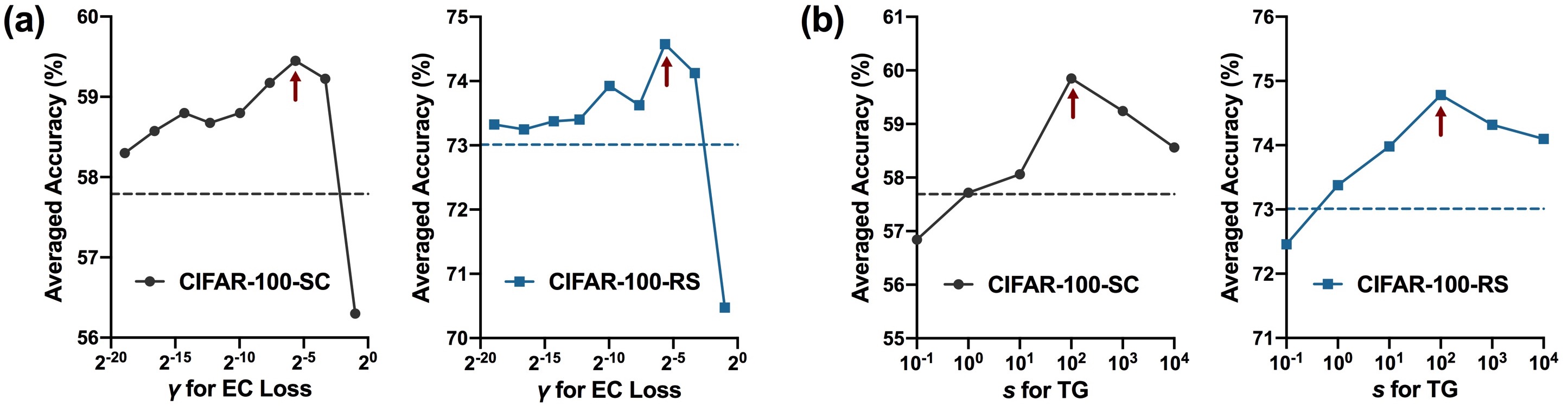} 
    \vspace{-.2cm}
	\caption{Effects of hyperparameters in CoSCL. (a) $\gamma$ for ensemble cooperation (EC) loss; (b) $s$ for task-adaptive gates (TG). The dashed lines indicate the performance w/o EC or TG in corresponding benchmarks. The arrows denote the chosen values.}
	\label{fig:hyperparameter}
    \vspace{-.5cm}
\end{figure}

\textbf{Detailed Analysis:}  Now, we use EWC \cite{kirkpatrick2017overcoming} as the default continual learning method and provide a detailed analysis for the superior performance of CoSCL. First, we analyze the \textbf{knowledge transfer} among tasks by evaluating the metrics of backward transfer (BWT), which is the averaged influence of learning each new task to the old tasks, and forward transfer (FWT), which is the averaged influence of remembering the old tasks to each new task \cite{lopez2017gradient}. As shown in Fig.~\ref{fig:knowledge_transfer}, CoSCL substantially improves both BWT and FWT of the default method, and in general far exceeds other representative baselines implemented in a single model. In particular, CoSCL raises BWT to almost zero, which means that catastrophic forgetting can be completely avoided.
We also evaluate the \textbf{expertise} of each continual learner across tasks  in Fig. \ref{fig:task_diversity} and Appendix C. The predictions made by each continual learner's representations differ significantly and complement with each other. The functional diversity can be naturally obtained from the randomness in architecture, such as the use of dropout and a different random initialization for each learner, and is explicitly regulated by our ensemble cooperation loss (Fig.~\ref{fig:hyperparameter}, a, discussed later).

\begin{table*}[t]
	\centering
	\caption{Ablation study. $A_t$: averaged accuracy (\%) of $t$ tasks learned so far. TG: task-adaptive gates; EC: ensemble cooperation loss. } 
    \vspace{-.3cm}
	\smallskip
	\resizebox{0.75\textwidth}{!}{ 
	\begin{tabular}{lccccccc}
		\specialrule{0.01em}{1.2pt}{1.5pt}
		 \multicolumn{2}{c}{} & \multicolumn{2}{c}{CIFAR-100-SC} & \multicolumn{2}{c}{CIFAR-100-RS} \\
       Methods& \#Param & \(A_{10}\) & \(A_{20}\) & \(A_{10}\) & \(A_{20}\) \\
       \specialrule{0.01em}{1.2pt}{1.7pt}
       Single Continual Learner &837K &52.25 \tiny{\(\pm 2.99 \)}&51.74 \tiny{\(\pm 1.74\)}&68.72 \tiny{\(\pm 0.24\)}&69.18 \tiny{\(\pm 0.69\)}\\
       Classifier Ensemble &901K &50.08 \tiny{\(\pm 1.65\)}&43.88 \tiny{\(\pm 0.79\)}&66.80 \tiny{\(\pm 1.45\)}&55.65 \tiny{\(\pm 0.32\)}\\
       \cdashline{1-6}[2pt/2pt]
       Feature Ensemble &773K &58.76 \tiny{\(\pm 3.72\)}&57.69 \tiny{\(\pm 1.42\)}&73.57 \tiny{\(\pm 0.50\)}&73.01 \tiny{\(\pm 1.22\)}\\ 
       Feature Ensemble + EC&773K &61.12 \tiny{\(\pm 3.11\)}&59.49 \tiny{\(\pm 1.59\)}&75.46 \tiny{\(\pm 1.35\)}&74.76 \tiny{\(\pm 0.84\)}\\ 
       Feature Ensemble + TG&799K &62.01 \tiny{\(\pm 3.36\)}&59.85 \tiny{\(\pm 1.77\)}&76.11 \tiny{\(\pm 0.98\)}&74.78 \tiny{\(\pm 0.41\)} \\ 
       \rowcolor{black!20}
       Feature Ensemble + EC + TG &799K &62.89 \tiny{\(\pm 3.05\)}&60.84 \tiny{\(\pm 0.95\)}&78.08 \tiny{\(\pm 1.25\)}&76.05 \tiny{\(\pm 0.65\)}\\
      \specialrule{0.01em}{1.2pt}{1.7pt}
	\end{tabular}
	}
	\label{table:ablation_study}
\end{table*}

Next, we present the results of an \textbf{ablation study} in Table~\ref{table:ablation_study}. We first consider a naive baseline that averages the predictions of multiple independently-trained small continual learners, referred to as the ``classifier ensemble (CE)''.
However, such a naive baseline even underperforms the single continual learner (SCL). In contrast, the proposed feature ensemble (FE) of multiple small continual learners can naturally achieve a superior performance, where the ensemble cooperation loss (EC) and the task-adaptive gates (TG) bring obvious benefits by properly adjusting for functional diversity among learners and exploiting the additional information from task labels, respectively. Then we evaluate the effect of \textbf{hyperparameters} in Fig.~\ref{fig:hyperparameter}. The hyperparameters of EC and TG are only moderately sensitive within a wide range. In this case, an appropriate (positive) strength of EC constrains the excessive diversity of predictions to improve the performance, while the continual learners will lose diversity if EC is too strong, resulting in a huge performance drop. If CoSCL cannot obtain sufficient diversity from the randomness of its architecture, the use of negative strength of EC can naturally serve this purpose, left for further work.

\begin{figure}[t]
	\centering
	\includegraphics[width=0.95\columnwidth]{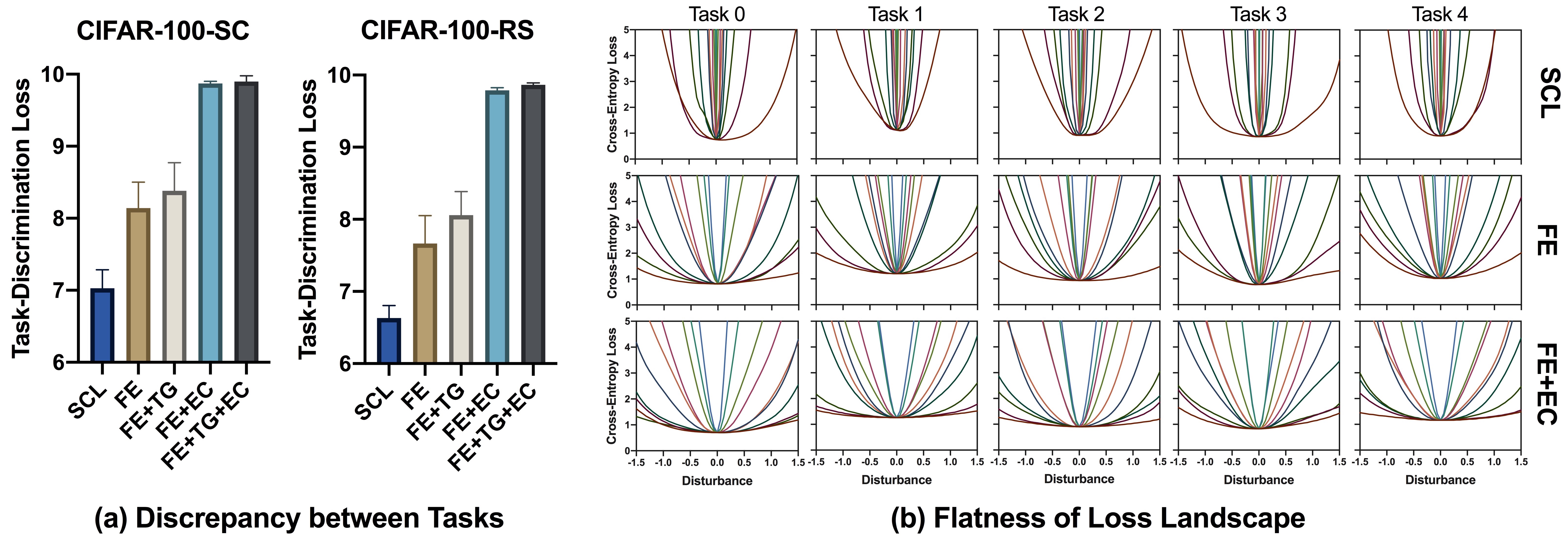} 
    \vspace{-.2cm}
	\caption{Empirical validation of our theoretical analysis. (a) Task-discrimination loss in feature space. Larger loss indicates a smaller $\mathcal{H}$-divergence. (b) Curvature of the test loss landscape for the first five incremental tasks on CIFAR-100-SC. Each line indicates the result of a random direction.}
	\label{fig:h_divergence_ec}
\end{figure}

Moreover, we empirically validate our theoretical analysis as below. We first evaluate the \textbf{$\mathcal{H}$-divergence} of feature representations between tasks, which relies on the capacity of a hypothesis space to distinguish them \cite{long2015learning}. Specifically, the $\mathcal{H}$-divergence can be empirically approximated by training a discriminator to distinguish if the features of input images belong to a task or not, where a larger discrimination loss indicates a smaller $\mathcal{H}$-divergence. As shown in Fig.~\ref{fig:h_divergence_ec}, a, the proposed FE together with EC can largely decrease the $\mathcal{H}$-divergence while TG has a moderate benefit (there is a saturation effect when they are combined together).
Meanwhile, we evaluate the \textbf{curvature of loss landscape} for the continually-learned solution by permuting the parameters to ten random directions \cite{deng2021flattening}, where the solution obtained by CoSCL enjoys a clearly flatter loss landscape than SCL (Fig.~\ref{fig:h_divergence_ec}, b).

Taking all results together, cooperating multiple small continual learners can mitigate the discrepancy between tasks in feature space and improve flatness of the continually-learned solution (Fig.~\ref{fig:h_divergence_ec}), thus facilitating both FWT and BWT (Fig.~\ref{fig:knowledge_transfer}). This is consistent with our theoretical analysis, suggesting that learning plasticity and memory stability are \emph{not} necessarily conflicting in continual learning, but can be simultaneously enhanced by a well-designed model architecture.

\begin{figure}[t]
	\centering
	\includegraphics[width=0.75\columnwidth]{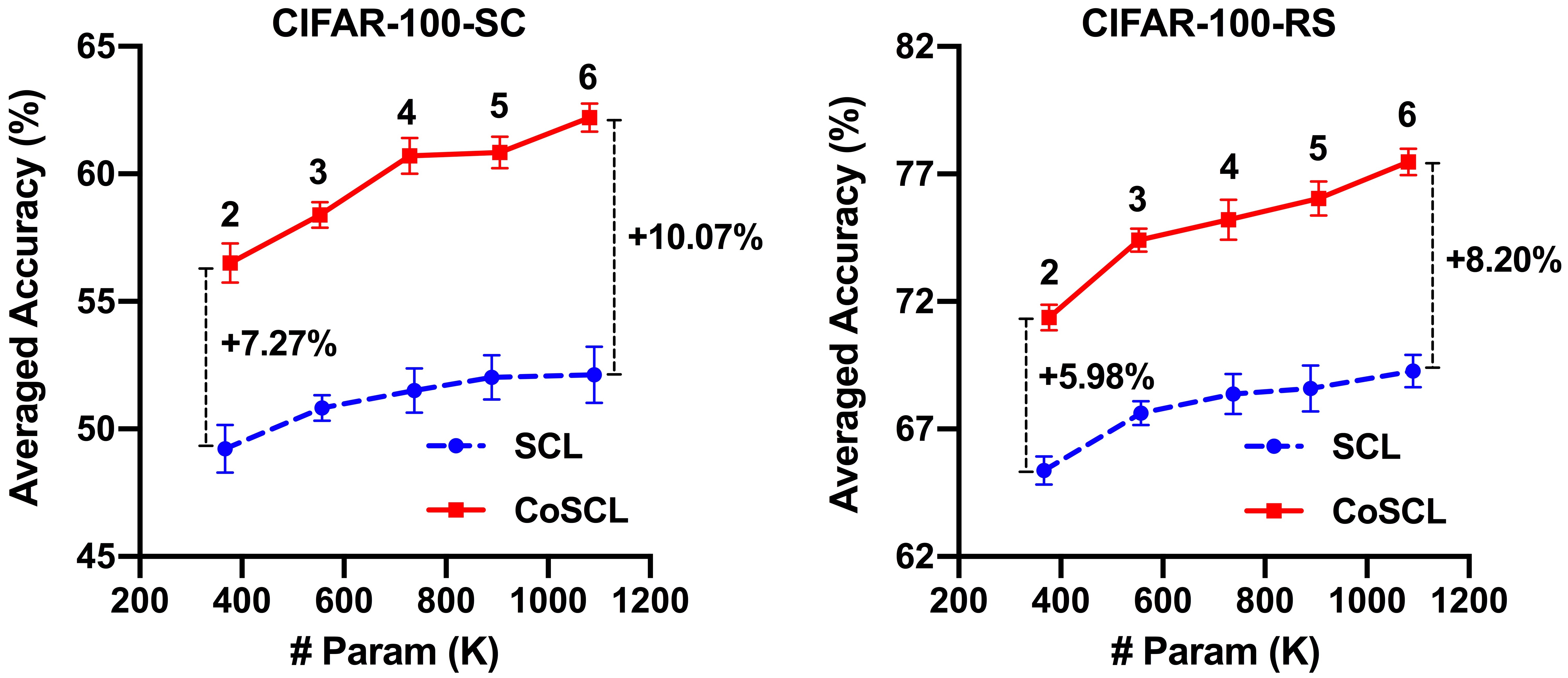}
    \vspace{-.2cm}
	\caption{Adding continual learners in CoSCL is more effective than widening the network of a single continual learner (SCL). We present the results of cooperating 2-6 continual learners with the same sub-network width, while accordingly adjust the size of SCL for a fair comparison.}
	\label{fig:scalability}
    \vspace{-.4cm}
\end{figure}

\textbf{Adding Continual Learners is More Effective than Widening a Single Network:} 
All of the above experiments are performed under a \emph{fixed} parameter budget. A recent work observed that a wider network usually suffers from less catastrophic forgetting \cite{mirzadeh2021wide}, providing an initial exploration of the effects of architecture in continual learning. Here we argue that adding continual learners with CoSCL is a better choice. In Fig.~\ref{fig:scalability} we compare the performance of using an increasing number of continual learners (the width is the same as that used in Table~\ref{table:visual_classification}) and accordingly widening the network of a single continual learner (SCL). It can be clearly seen that the performance gap between CoSCL and SCL is growing when more parameters are used. Therefore, CoSCL presents a promising direction for continual learning that can leverage network parameters in an efficient and scalable way.

\begin{wraptable}{r}{0.49\textwidth}
	\centering
    \vspace{-0.2cm}
	\caption{Averaged accuracy (\%) of unsupervised continual learning on CIFAR-100-RS. The results are reproduced from the officially-released code of \cite{madaan2021rethinking}.} 
	\smallskip
	\resizebox{0.49\textwidth}{!}{ 
	\begin{tabular}{lcc}
		\specialrule{0.01em}{1.2pt}{1.5pt}
       Methods & SimSiam \cite{chen2021exploring}& BarlowTwins \cite{zbontar2021barlow} \\
       \specialrule{0.01em}{1.2pt}{1.7pt}
       Finetune &  41.38 \tiny{\(\pm 0.80\)}& 63.29 \tiny{\(\pm 0.38\)}\\        
       \rowcolor{black!20}
      \ \emph{w/} CoSCL &\textbf{46.33} \tiny{\(\pm 0.51\)}  &\textbf{74.03} \tiny{\(\pm 0.36\)}\\
      \specialrule{0.01em}{1.2pt}{1.7pt}
	\end{tabular}
	}
	\label{table:ucl}
	\vspace{-.5cm}
\end{wraptable}

\textbf{Unsupervised Continual Learning (UCL):} has the unique property of being naturally robust to catastrophic forgetting when fine-tuning on incremental unlabeled data \cite{hu2021well,madaan2021rethinking}. An empirical explanation is that UCL achieves a flatter loss landscape and more meaningful feature representations \cite{madaan2021rethinking}, which is consistent with our analysis. We further validate this idea by adapting CoSCL to UCL\footnote{Here we only use feature ensemble (FE) with ensemble cooperation loss (EC).}, where we follow the UCL setting of \cite{madaan2021rethinking} for CIFAR-100-RS and use a similar architecture as Table~\ref{table:visual_classification}. As shown in Table~\ref{table:ucl}, CoSCL can significantly improve the performance of UCL with two strong unsupervised learning strategies such as SimSiam \cite{chen2021exploring} and BarlowTwins \cite{zbontar2021barlow}.

\section{Conclusion}

Numerous efforts in continual learning have been devoted to developing effective approaches based on a single model, but their efficacy might be limited by such a priori assumption. 
In this work, we present a unified form of generalization bounds for learning plasticity and memory stability in continual learning, consisting of three components, and demonstrate that the both aspects can be simultaneously improved by a compatible parameter space of a well-designed mode architecture.
Inspired by the robust biological learning system, we propose to cooperate multiple (small) continual learners, which can naturally tighten the generalization bounds through improving the three components. Our method can substantially enhance the performance of representative continual learning strategies by improving both learning plasticity and memory stability. We hope that this work can serve as a strong baseline to stimulate new ideas for continual learning from an architecture perspective. A promising direction is to cooperate a variety of continual learning approaches with properly-designed architectures, so as to fully leverage task attributes for desired compatibility.

~\\
\textbf{Acknowledgements.}
This work was supported by the National Key Research and Development Program of China (2017YFA0700904, 2020AAA0106000, \\2020AAA0104304, 2020AAA0106302, 2021YFB2701000), NSFC Projects (Nos. 62061136001, 62106123, 62076147, U19B2034, U1811461, U19A2081, 61972224), Beijing NSF Project (No. JQ19016), BNRist (BNR2022RC01006), Tsinghua-Peking Center for Life Sciences, Tsinghua Institute for Guo Qiang, Beijing Academy of Artificial Intelligence (BAAI), Tsinghua-OPPO Joint Research Center for Future Terminal Technology, the High Performance Computing Center, Tsinghua University, and China Postdoctoral Science Foundation (No. 2021T140377, 2021M701892).


\clearpage

%
%
\bibliographystyle{splncs04}
\bibliography{main}

\clearpage

\appendix

\section{Complete Proof of Theoretical Analysis}


\subsection{Proof of Proposition 1}

We assume that a distribution $\mathbb{D}$ is with input space $\mathcal{X}$ and a global label function $h: \mathcal{X} \to \mathcal{Y}$, where $\mathcal{Y}$ denotes a label space, and $h(x)$ generates target label for all the input, i.e., $y = h(x)$.
Consider a bounded loss function $\ell: \mathcal{Y} \times \mathcal{Y} \to [0, c]$ (where $c$ is the upper bound), such that $\ell(y_1, y_2)=0$ holds if and only if $y_1=y_2$. 
Then, we define a population loss over the distribution $\mathbb{D}$ by
$\mathcal{E}_{\mathbb{D}} (\theta) = \mathcal{E}_{\mathbb{D}} (f_{\theta}, h) := \mathbb{E}_{(x, y) \sim \mathbb{D}}[\ell(f_{\theta}(x), h(x))]$.
Let $D$ denote a training set following the distribution $\mathbb{D}$ with $N$ data-label pairs.
To minimize $\mathcal{E}_{\mathbb{D}} (\theta)$, we can minimize an empirical risk over the training set $D$ in a parameter space, i.e., $ \min_{\theta} \hat{\mathcal{E}}_{D} (\theta)$.
Further, to find a flat solution, we define a robust empirical risk by the worst case of the neighborhood in parameter space as $\hat{\mathcal{E}}_{D}^b (\theta) := \rm{max}_{\lVert \Delta \rVert \leq b} \hat{\mathcal{E}}_{D} (\theta +\Delta)$, where $b$ is the radius around $\theta$ and $\lVert \cdot \rVert$ denotes the L2 norm.

Below are one important definition and three critical lemmas for the proof of Proposition~\ref{errorbound}.

\begin{definition}
(Based on Definition 1 of \cite{ben2010theory})
Given two distributions, $\mathbb{T}$ and $\mathbb{S}$, let $\mathcal{H}$ be a hypothesis class on input space $\mathcal{X}$ and denote by $I(h)$ the set for which $h \in \mathcal{H}$ is the characteristic function: that is, $x \in  I(h) \Leftrightarrow  h(x)=1$. The $\mathcal{H}$-divergence between $\mathbb{T}$ and $\mathbb{S}$ is
\begin{align}
    \rm{Div}(\mathbb{T}, \mathbb{S}) = 2 \sup_{\textit{h} \in \mathcal{H}} |\mathcal{P}_{ \mathbb{T}}(\textit{I}(\textit{h})) - \mathcal{P}_{ \mathbb{S}}(\textit{I}(\textit{h}))|.
\end{align}
\end{definition}

\begin{lemma} \label{lemmaDIV}
Let $\mathbb{S} = \{\mathbb{S}_i\}_{i=1}^{s}$ and $\mathbb{T}$ be $s$ source distributions and the target distribution, respectively. The $\mathcal{H}$-divergence between $\{\mathbb{S}_i\}_{i=1}^{s}$ and $\mathbb{T}$ is bounded as follows:
\begin{align}
    \rm{Div}(\mathbb{S}, \mathbb{T}) \leq \frac{1}{s} \sum_{\textit{i}=1}^{\textit{s}} \rm{Div}(\mathbb{S}_i, \mathbb{T}).
\end{align}
\end{lemma}
\noindent\emph{Proof.}
By the definition of $\mathcal{H}$-divergence,
\begin{align}
\begin{split}
    \rm{Div}(\mathbb{S}, \mathbb{T}) & = 2 \sup_{\textit{h} \in \mathcal{H}} |\mathcal{P}_{ \mathbb{S}}(\textit{I}(\textit{h})) - \mathcal{P}_{ \mathbb{T}}(\textit{I}(\textit{h}))| \\
    & =2  \sup_{\textit{h} \in \mathcal{H}} \left|\sum_{i=1}^{s} \frac{1}{s}(\mathcal{P}_{ \mathbb{S}_i}(\textit{I}(\textit{h})) - \mathcal{P}_{ \mathbb{T}}(\textit{I}(\textit{h}))) \right| \\
    & \leq 2 \sup_{\textit{h} \in \mathcal{H}} \sum_{i=1}^{s} \frac{1}{s}  \left|   \mathcal{P}_{ \mathbb{S}_i}(\textit{I}(\textit{h})) - \mathcal{P}_{ \mathbb{T}}(\textit{I}(\textit{h})) \right| \\
    &  \leq 2  \sum_{i=1}^{s}  \frac{1}{s} \sup_{\textit{h} \in \mathcal{H}} \left|   \mathcal{P}_{ \mathbb{S}_i}(\textit{I}(\textit{h})) - \mathcal{P}_{ \mathbb{T}}(\textit{I}(\textit{h})) \right| \\       
    & =  \frac{1}{s} \sum_{i=1}^{s} \rm{Div}(\mathbb{S}_i, \mathbb{T}),
    \end{split}
\end{align}
where the first inequality is due to the triangle inequality and the second inequality is by the additivity of the $\sup$ function. This finishes the proof.

\begin{lemma} \label{lemma1} 
Given two distributions, $\mathbb{T}$ and $\mathbb{S}$, the difference between the population loss with $\mathbb{T}$ and $\mathbb{S}$ is bounded by the divergence between $\mathbb{T}$ and $\mathbb{S}$ as follows:
\begin{align}
|\mathcal{E}_{\mathbb{T}} (f_1, h_1) - \mathcal{E}_{\mathbb{S}} (f_1, h_1)| \leq \frac{1}{2}\rm{Div}(\mathbb{T}, \mathbb{S}),
\end{align}
where $\rm{Div}(\mathbb{T}, \mathbb{S}) := 2 \sup_{\textit{h} \in \mathcal{H}} |\mathcal{P}_{ \mathbb{T}}(\textit{I}(\textit{h})) - \mathcal{P}_{ \mathbb{S}}(\textit{I}(\textit{h}))|$ is the $\mathcal{H}$-divergence for the distribution $\mathbb{T}$ and $\mathbb{S}$ ($I(h)$ is the characteristic function).
\end{lemma}

\noindent\emph{Proof.}
By the definition of $\mathcal{H}$-divergence,
\begin{align}
\begin{split}
    \rm{Div}(\mathbb{T}, \mathbb{S}) & = 2 \sup_{\textit{h} \in \mathcal{H}} |\mathcal{P}_{ \mathbb{T}}(\textit{I}(\textit{h})) - \mathcal{P}_{ \mathbb{S}}(\textit{I}(\textit{h}))| \\
   &  = 2 \sup_{f_1,h_1 \in \mathcal{H}} \left| \mathcal{P}_{(x, y) \sim \mathbb{T}} [f_1(x) \neq h_1(x)] - \mathcal{P}_{(x, y) \sim \mathbb{S}} [f_1(x) \neq h_1(x)] \right| \\
   &= 2 \sup_{f_1,h_1 \in \mathcal{H}}  \left| \mathbb{E}_{(x, y) \sim \mathbb{T}}[\ell(f_1(x), h_1(x))] - \mathbb{E}_{(x, y) \sim \mathbb{S}}[\ell(f_1(x), h_1(x))] \right|\\
   & =2 \sup_{f_1,h_1 \in \mathcal{H}} |\mathcal{E}_{\mathbb{T}} (f_1, h_1) - \mathcal{E}_{\mathbb{S}} (f_1, h_1)| \\
   & \geq 2 |\mathcal{E}_{\mathbb{T}} (f_1, h_1) - \mathcal{E}_{\mathbb{S}} (f_1, h_1)|.
    \end{split}
\end{align}
It completes the proof.

\begin{lemma}\label{lemma2}
Let $\Theta$ be a cover of a parameter space with VC dimension $d$.
Then, for any $\delta \in (0,1)$ with probability at least $1-\delta$,
for any $\theta \in \Theta$:
\begin{align}               
   | \mathcal{E}_{\mathbb{D}} (\theta) - \hat{\mathcal{E}}_{D}^b (\theta) |  \leq \sqrt{\frac{{\textit d} [\ln (\textit N /\textit d) ] + \ln (1/\delta)}{2 \textit N} },
\end{align}
where $ \hat{\mathcal{E}}_{D}^b (\theta)  $ is a robust empirical risk with $N$ samples in its training set $D$, and $b$ is the radius around $\theta$.
\end{lemma}

\noindent\emph{Proof.}
For the distribution $\mathbb{D}$, we have
\begin{align}
    \mathcal{P} (|\mathcal{E}_{\mathbb{D}} (\theta) - \hat{\mathcal{E}}_{D} (\theta)| \geq \epsilon) \leq 2  m_{{\Theta}}(N) \exp(-2N\epsilon^{2}) ,
\end{align}
where $m_{{\Theta}}(N)$ is the amount of all possible prediction results for $N$ samples, which implies the model complexity in the parameter space $\Theta$. We set $m_{{\Theta}}(N) = \frac{1}{2} \left(\frac{ N}{d}\right)^d$ in our model, and assume a confidence bound $\epsilon = \sqrt{\frac{{\textit d} [\ln (\textit N /\textit d) ] + \ln (1/\delta)}{2 \textit N} }$. Then we get
\begin{align}
    \mathcal{P} (|\mathcal{E}_{\mathbb{D}} (\theta) - \hat{\mathcal{E}}_{D} (\theta)| \geq \epsilon) \leq \left(\frac{ N}{d}\right)^d\exp(-2N\epsilon^{2}) = \delta.
\end{align}
Hence, the inequality $|\mathcal{E}_{\mathbb{D}} (\theta) - \hat{\mathcal{E}}_{D} (\theta) |  \leq \epsilon $ holds with probability at least $1-\delta$. Further, based on the fact that $\hat{\mathcal{E}}_{D}^b (\theta) \geq \hat{\mathcal{E}}_{D} (\theta)$, we have
\begin{align}
    |\mathcal{E}_{\mathbb{D}} (\theta) - \hat{\mathcal{E}}_{D}^b (\theta) | \leq |\mathcal{E}_{\mathbb{D}} (\theta) - \hat{\mathcal{E}}_{D} (\theta) | \leq \epsilon .
\end{align}
It completes the proof.


\noindent \textbf{Proof of Proposition~\ref{errorbound}} 
If we continually learn $t$ tasks that follow the distribution $\mathbb{D}_1,\cdots,\mathbb{D}_t$, then a solution $\theta_{1:t}$ can be obtained.
In addition, let $\theta_{t}$ denote a solution obtained over the distribution $\mathbb{D}_t$ only, and $\theta_{1:t-1}$ be a solution obtained over the set of distribution $\mathbb{D}_1,\cdots,\mathbb{D}_{t-1}$.
Then, we have
\begin{small}
\begin{align}
\begin{split}
 & \mathcal{E}_{\mathbb{D}_t} (\theta_{1:t-1})  \leq \mathcal{E}_{\mathbb{D}_{1:t-1}} (\theta_{1:t-1}) + \frac{1}{2} \rm{Div}(\mathbb{D}_{1:\textit{t}-1}, \mathbb{D}_\textit{t}) \\
 & \leq \hat{\mathcal{E}}_{D_{1:t-1}}^b (\theta_{1:t-1}) + \frac{1}{2} \rm{Div}(\mathbb{D}_{1:\textit{t}-1}, \mathbb{D}_\textit{t}) + \sqrt{\frac{{\textit d}[\ln (\textit N_{1:\textit t-1}/\textit d) ] + \ln (1/\delta)}{2 \textit N_{1:\textit t-1}} } \\
 & \leq \hat{\mathcal{E}}_{D_{1:t-1}}^b (\theta_{1:t-1}) +\frac{1}{2(t-1)}\sum_{k=1}^{t-1} \rm{Div}(\mathbb{D}_\textit{k}, \mathbb{D}_\textit{t}) + \sqrt{\frac{{\textit d}[\ln (\textit N_{1:\textit t-1}/\textit d) ] + \ln (1/\delta)}{2 \textit N_{1:\textit t-1}} } \\
 & \leq \hat{\mathcal{E}}_{D_{1:t-1}}^b (\theta_{1:t-1}) +\frac{1}{2(t-1)}\sum_{k=1}^{t-1} \rm{Div}(\mathbb{D}_\textit{k}, \mathbb{D}_\textit{t}) + \sqrt{\frac{{\textit d}[\ln (\textit N_{1:\textit t-1}/\textit d) ] + \ln (1/\delta)}{ \textit N_{1:\textit t-1}} },
 \end{split}
\end{align}
\end{small}
where the first three inequalities are from Lemma~\ref{lemma1}, Lemma~\ref{lemma2} and Lemma~\ref{lemmaDIV}, respectively. $\mathbb{D}_{1:t-1}:=\{\mathbb{D}_k\}_{k=1}^{t-1}$ and we rewrite a mixture of all the $t-1$ distributions as $\mathbb{D}_{1:t-1}:= \frac{1}{t-1} \sum_{k=1}^{t-1} \mathbb{D}_k$ using convex combination. $N_{1:t-1}=\sum_{k=1}^{t-1} N_k$ is the total number of training samples over all $t-1$ old tasks.

Further, we have
\begin{small}
\begin{align}
\begin{split}
& \mathcal{E}_{\mathbb{D}_t} (\theta_{1:t})   < \mathcal{E}_{\mathbb{D}_t} (\theta_{1:t-1})  \\
& \leq \hat{\mathcal{E}}_{D_{1:t-1}}^b (\theta_{1:t-1}) +\frac{1}{2(t-1)}\sum_{k=1}^{t-1} \rm{Div}(\mathbb{D}_\textit{k}, \mathbb{D}_\textit{t}) + \sqrt{\frac{{\textit d}[\ln (\textit N_{1:\textit t-1}/\textit d) ] + \ln (1/\delta)}{ \textit N_{1:\textit t-1}} }  \\
& \leq \hat{\mathcal{E}}_{D_{1:t-1}}^b (\theta_{1:t}) +\frac{1}{2(t-1)}\sum_{k=1}^{t-1} \rm{Div}(\mathbb{D}_\textit{k}, \mathbb{D}_\textit{t}) + \sqrt{\frac{{\textit d}[\ln (\textit N_{1:\textit t-1}/\textit d) ] + \ln (1/\delta)}{ \textit N_{1:\textit t-1}} }.
\end{split}
\end{align}
\end{small}
Likewise, we get
\begin{align}
\begin{split}
 & \mathcal{E}_{\mathbb{D}_{1:t-1}} (\theta_{t})  \leq \mathcal{E}_{\mathbb{D}_{t}} (\theta_{t}) + \frac{1}{2} \rm{Div}(\mathbb{D}_{\textit{t}}, \mathbb{D}_{1:\textit{t}-1}) \\
 & \leq \hat{\mathcal{E}}_{D_{t}}^b (\theta_{t}) + \frac{1}{2} \rm{Div}(\mathbb{D}_\textit{t}, \mathbb{D}_{1:\textit{t}-1}) + \sqrt{\frac{{\textit d}[\ln (\textit N_{\textit t}/\textit d) ] + \ln (1/\delta)}{2  \textit N_{\textit t}}  } \\
 & \leq \hat{\mathcal{E}}_{D_{t}}^b (\theta_{t})  +\frac{1}{2(t-1)}\sum_{k=1}^{t-1} \rm{Div}(\mathbb{D}_\textit{t}, \mathbb{D}_{\textit{k}}) + \sqrt{\frac{{\textit d}[\ln (\textit N_{\textit t}/\textit d) ] + \ln (1/\delta)}{2 \textit N_{\textit t}} }\\
 & \leq \hat{\mathcal{E}}_{D_{t}}^b (\theta_{t})  +\frac{1}{2(t-1)}\sum_{k=1}^{t-1} \rm{Div}(\mathbb{D}_\textit{t}, \mathbb{D}_{\textit{k}}) + \sqrt{\frac{{\textit d}[\ln (\textit N_{\textit t}/\textit d) ] + \ln (1/\delta)}{ \textit N_{\textit t}} }.
 \end{split}
\end{align}

Further, we have
\begin{align}
\begin{split}
 & \mathcal{E}_{\mathbb{D}_{1:t-1}} (\theta_{1:t})  < \mathcal{E}_{\mathbb{D}_{1:t-1}} (\theta_{t}) \\
 & \leq   \hat{\mathcal{E}}_{D_{t}}^b (\theta_{t})  +\frac{1}{2(t-1)}\sum_{k=1}^{t-1} \rm{Div}(\mathbb{D}_\textit{t}, \mathbb{D}_{\textit{k}}) + \sqrt{\frac{{\textit d}[\ln (\textit N_{\textit t}/\textit d) ] + \ln (1/\delta)}{ \textit N_{\textit t}} } \\
 & \leq  \hat{\mathcal{E}}_{D_{t}}^b (\theta_{1:t}) + \frac{1}{2(t-1)}\sum_{k=1}^{t-1} \rm{Div}(\mathbb{D}_\textit{t}, \mathbb{D}_{\textit{k}}) + \sqrt{\frac{{\textit d}[\ln (\textit N_{\textit t}/\textit d) ] + \ln (1/\delta)}{ \textit N_{\textit t}} } ,
\end{split}
\end{align}
where $N_t$ is the number of training samples over the distribution $\mathbb{D}_t$.

Combining all the inequalities above finishes the proof.

\subsection{Proof of Proposition 2}
\noindent \textbf{Proof of Proposition~\ref{gapbound}} 
Let $\hat \theta_{1:t}^{b}$ denote the optimal solution of the continually learned $1:t$ tasks by
robust empirical risk minimization over the new task, i.e., $\hat \theta_{1:t}^{b} = \arg \min_{\theta \in \Theta} \hat{\mathcal{E}}_{D_t}^b (\theta)$, where $\Theta$ denotes a cover of a parameter space with VC dimension $d$.
Likewise, let $\hat \theta_{t}^{b}$ be the optimal solution by robust empirical risk minimization over the distribution $\mathbb{D}_t$ only, and $\hat \theta_{1:t-1}^{b}$ over the set of distribution $\mathbb{D}_1,\cdots,\mathbb{D}_{t-1}$. That is, $\hat \theta_{t}^{b} = \arg \min_{\theta} \hat{\mathcal{E}}_{D_t}^b (\theta)$ and $\hat \theta_{1:t-1}^{b} = \arg \min_{\theta} \hat{\mathcal{E}}_{D_{1:t-1}}^b (\theta)$.

Then, let $ \theta_{t}$ be the optimal solution over the distribution $\mathbb{D}_t$ only, i.e., $\theta_{t} = \arg \min_{\theta} {\mathcal{E}}_{\mathbb{D}_t} (\theta)$. From Lemma~\ref{lemma2}, the following inequality holds with probability at least $1-\frac{\delta}{2}$,
\begin{align} \label{pro2-Dt-1}
\begin{split}
    |\mathcal{E}_{\mathbb{D}_{1:t-1}} (\theta_{t}) - \hat{\mathcal{E}}_{D_{1:t-1}} (\theta_{t})| & \leq \sqrt{\frac{d\ln (N_{1:t-1}/d) + \ln (2/\delta)}{2N_{1:t-1}} } \\
   & \leq \sqrt{\frac{d\ln (N_{1:t-1}/d) + \ln (2/\delta)}{N_{1:t-1}} },
    \end{split}
\end{align}
where $N_{1:t-1}=\sum_{k=1}^{t-1} N_k$ is the total number of training samples over all $t-1$ old tasks. Then, we have
\begin{small}
\begin{align} \label{pro-dd}
\begin{split}
    \min_{\theta \in \Theta} \hat{\mathcal{E}}_{D_{1:t-1}} (\theta) 
    & \leq \hat{\mathcal{E}}_{D_{1:t-1}} (\theta_{t}) \\ 
    & \leq {\mathcal{E}}_{\mathbb{D}_{1:t-1}}(\theta_{t}) + \sqrt{\frac{\textit{d}\ln (N_{1:t-1}/\textit{d}) + \ln (2/\delta)}{N_{1:t-1}} } \\
    & \leq {\mathcal{E}}_{\mathbb{D}_{t}}(\theta_{t}) + \frac{1}{2} \rm{Div}(\mathbb{D}_{1:\textit{t}-1}, \mathbb{D}_\textit{t}) + \sqrt{\frac{\textit{d}\ln (\textit{N}_{1:\textit{t}-1}/\textit{d}) + \ln (2/\delta)}{\textit{N}_{1:\textit{t}-1}} } \\
    & = \min_{\theta \in \Theta} \mathcal{E}_{\mathbb{D}_{t}} (\theta )  + \frac{1}{2} \rm{Div}(\mathbb{D}_{1:\textit{t}-1}, \mathbb{D}_\textit{t}) + \sqrt{\frac{\textit{d}\ln (\textit{N}_{1:\textit{t}-1}/\textit{d}) + \ln (2/\delta)}{\textit{N}_{1:\textit{t}-1}} } \\
    & \leq \min_{\theta \in \Theta} \mathcal{E}_{\mathbb{D}_{t}} (\theta )  + \frac{1}{2(t-1)}\sum_{k=1}^{t-1} \rm{Div}(\mathbb{D}_\textit{k}, \mathbb{D}_{\textit{t}})   + \sqrt{\frac{\textit{d}\ln (\textit{N}_{1:\textit{t}-1}/\textit{d}) + \ln (2/\delta)}{\textit{N}_{1:\textit{t}-1}} },
    \end{split} 
\end{align}
\end{small}
where the third inequality holds from Lemma~\ref{lemma1}, and the final inequality is from Lemma~\ref{lemmaDIV}.

From Proposition~\ref{errorbound}, the following inequality holds with probability at least $1-\frac{\delta}{2}$,
\begin{small}
\begin{align} \label{pro2-edt} 
     \mathcal{E}_{\mathbb{D}_t} (\hat \theta_{1:t-1}^{b}) <  \hat{\mathcal{E}}_{D_{1:t-1}}^b (\hat \theta_{1:t-1}^{b}) + \frac{1}{2(t-1)}\sum_{k=1}^{t-1} \rm{Div}(\mathbb{D}_\textit{k}, \mathbb{D}_\textit{t}) + \sqrt{\frac{{\textit d}\ln (\textit N_{1:\textit t-1}/\textit d) + \ln (2/\delta)}{\textit N_{1:\textit t-1}} }.
\end{align}
\end{small}
Combining Eqn.~\ref{pro-dd} and Eqn.~\ref{pro2-edt}, we get
\begin{small}
\begin{align} \label{pro2-part1}
    \begin{split}
      & \mathcal{E}_{\mathbb{D}_{t}} (\hat \theta_{1:t}^{b}) - \min_{\theta \in \Theta} \mathcal{E}_{\mathbb{D}_{t}} (\theta )  \leq \mathcal{E}_{\mathbb{D}_{t}} (\hat \theta_{1:t-1}^{b}) - \min_{\theta \in \Theta} \mathcal{E}_{\mathbb{D}_{t}} (\theta ) \\
       &  \leq \hat{\mathcal{E}}_{D_{1:t-1}}^b (\hat \theta_{1:t-1}^{b}) - \min_{\theta \in \Theta} \hat{\mathcal{E}}_{D_{1:t-1}} (\theta)  +\frac{1}{t-1}\sum_{k=1}^{t-1} \rm{Div}(\mathbb{D}_\textit{k}, \mathbb{D}_\textit{t}) + 2 \sqrt{\frac{{\textit d}\ln (\textit N_{1:\textit t-1}/\textit d) + \ln (2/\delta)}{\textit N_{1:\textit t-1}} } \\
       &=\min_{\theta \in \Theta} \hat{\mathcal{E}}_{D_{1:t-1}}^b (\theta) - \min_{\theta \in \Theta} \hat{\mathcal{E}}_{D_{1:t-1}} (\theta)  +\frac{1}{t-1}\sum_{k=1}^{t-1} \rm{Div}(\mathbb{D}_\textit{k}, \mathbb{D}_\textit{t}) + 2 \sqrt{\frac{{\textit d}\ln (\textit N_{1:\textit t-1}/\textit d) + \ln (2/\delta)}{\textit N_{1:\textit t-1}} }.
    \end{split}
\end{align}
\end{small}
This completes the first part of Proposition~\ref{gapbound}.

Similarly, let $ \theta_{1:t}$ be the optimal solution over the distribution $\mathbb{D}_{1:t-1}$ only, i.e., $\theta_{1:t-1} = \arg \min_{\theta} {\mathcal{E}}_{\mathbb{D}_{1:t-1}} (\theta)$. From Lemma~\ref{lemma2}, the following inequality holds with probability at least $1-\frac{\delta}{2}$,
\begin{align} \label{pro2-Dt-2}
\begin{split}
    |\mathcal{E}_{\mathbb{D}_{t}} (\theta_{1:t-1}) - \hat{\mathcal{E}}_{D_{t}} (\theta_{1:t-1})|   &  \leq \sqrt{\frac{d\ln (N_{t}/d) + \ln (2/\delta)}{2N_{t}} }\\
    & \leq \sqrt{\frac{d\ln (N_{t}/d) + \ln (2/\delta)}{N_{t}} },
\end{split}
\end{align}
where $N_t$ is the number of training samples in the distribution $\mathbb{D}_{t}$. Then, we have 
\begin{align} \label{pro-dd-2}
\begin{split}
    \min_{\theta \in \Theta} \hat{\mathcal{E}}_{D_{t}} (\theta) 
    & \leq \hat{\mathcal{E}}_{D_{t}} (\theta_{1:t-1}) \\ 
    & \leq {\mathcal{E}}_{\mathbb{D}_{t}}(\theta_{1:t-1}) + \sqrt{\frac{d\ln (N_{t}/d) + \ln (2/\delta)}{N_{t}} } \\
    & \leq {\mathcal{E}}_{\mathbb{D}_{1:t-1}}(\theta_{1:t-1}) + \frac{1}{2} \rm{Div}(\mathbb{D}_{\textit{t}}, \mathbb{D}_{1:\textit{t}-1}) + \sqrt{\frac{\textit{d}\ln (\textit{N}_{\textit{t}}/\textit{d}) + \ln (2/\delta)}{\textit{N}_{\textit{t}}} } \\
    & = \min_{\theta \in \Theta} \mathcal{E}_{\mathbb{D}_{1:t-1}} (\theta )  + \frac{1}{2} \rm{Div}(\mathbb{D}_{\textit{t}}, \mathbb{D}_{1:\textit{t}-1}) + \sqrt{\frac{\textit{d}\ln (\textit{N}_{\textit{t}}/\textit{d}) + \ln (2/\delta)}{\textit{N}_{\textit{t}}} } \\
    & \leq \min_{\theta \in \Theta} \mathcal{E}_{\mathbb{D}_{1:t-1}} (\theta )  + \frac{1}{2(t-1)}\sum_{k=1}^{t-1} \rm{Div}(\mathbb{D}_\textit{t}, \mathbb{D}_{\textit{k}})   + \sqrt{\frac{\textit{d}\ln (\textit{N}_{\textit{t}}/\textit{d}) + \ln (2/\delta)}{\textit{N}_{\textit{t}}} },
    \end{split} 
\end{align}
where the third inequality holds from Lemma~\ref{lemma1}, and the final inequality is from Lemma~\ref{lemmaDIV}.
From Proposition~\ref{errorbound}, the following inequality holds with probability at least $1-\frac{\delta}{2}$,
\begin{align} \label{pro2-edt-2} 
     \mathcal{E}_{\mathbb{D}_{1:t-1}} (\hat \theta_{t}^{b}) <  \hat{\mathcal{E}}_{D_{t}}^b (\hat \theta_{t}^{b}) + \frac{1}{2(t-1)}\sum_{k=1}^{t-1} \rm{Div}(\mathbb{D}_\textit{t}, \mathbb{D}_\textit{k}) + \sqrt{\frac{{\textit d}\ln (\textit N_{\textit t}/\textit d) + \ln (2/\delta)}{\textit N_{\textit t}} }.
\end{align}

Combining Eqn.~\ref{pro-dd-2} and Eqn.~\ref{pro2-edt-2}, we get
\begin{align} \label{pro2-part2}
    \begin{split}
      & \mathcal{E}_{\mathbb{D}_{1:t-1}} (\hat \theta_{1:t}^{b}) - \min_{\theta \in \Theta} \mathcal{E}_{\mathbb{D}_{1:t-1}} (\theta )  \leq  \mathcal{E}_{\mathbb{D}_{1:t-1}} (\hat \theta_{t}^{b}) - \min_{\theta \in \Theta} \mathcal{E}_{\mathbb{D}_{1:t-1}} (\theta )  \\
       &  \leq \hat{\mathcal{E}}_{D_{t}}^b (\hat \theta_{t}^{b})  -\min_{\theta \in \Theta} \hat{\mathcal{E}}_{D_{t}} (\theta)  +\frac{1}{t-1}\sum_{k=1}^{t-1} \rm{Div}(\mathbb{D}_\textit{t}, \mathbb{D}_\textit{k}) + 2 \sqrt{\frac{{\textit d}\ln (\textit N_{\textit t}/\textit d) + \ln (2/\delta)}{\textit N_{\textit t}} } \\
       &=\min_{\theta \in \Theta} \hat{\mathcal{E}}_{D_{t}}^b (\theta) - \min_{\theta \in \Theta} \hat{\mathcal{E}}_{D_{t}} (\theta)  +\frac{1}{t-1}\sum_{k=1}^{t-1} \rm{Div}(\mathbb{D}_\textit{t}, \mathbb{D}_\textit{k}) + 2 \sqrt{\frac{{\textit d}\ln (\textit N_{\textit t}/\textit d) + \ln (2/\delta)}{\textit N_{\textit t}} }. 
    \end{split}
\end{align}
This completes the second part of Proposition~\ref{gapbound}.

\subsection{Proof of Proposition~\ref{coscl}}

Below is one critical lemma for the proof of Proposition~\ref{coscl}.

\begin{lemma}\label{pro3-lemma}
Let $\{\Theta_i \in  \mathbb{R}^{r}\}_{i=1}^{K}$ be a set of $K$ parameter spaces ($K>1$ in general), 
$d_i$ be a VC dimension of $\Theta_i$, and $\Theta = \cup_{i=1}^{K}\Theta_i$ with VC dimension $d$.
Let $\theta_i = \arg \max_{\theta \in \Theta_i} \mathcal{E}_{\mathbb{D}} (\theta)$ be a local maximum in the $i$-th parameter space (i.e., $i$-th ball). 
Then, for any $\delta \in (0,1)$ with probability at least $1-\delta$,
for any $\theta \in \Theta$:
\begin{align}               
   | \mathcal{E}_{\mathbb{D}} (\theta) - \hat{\mathcal{E}}_{D}^b (\theta) |  \leq 
   \max_{i \in [1,K]} \sqrt{\frac{d_i\ln (N/d_i) + \ln (K/\delta)}{2N} },
\end{align}
where $ \hat{\mathcal{E}}_{D}^b (\theta)  $ is a robust empirical risk with $N$ samples in its training set $D$, and $b$ is the radius around $\theta$.
\end{lemma}

\noindent\emph{Proof.}
For the distribution $\mathbb{D}$, we have
\begin{align}
    \begin{split}
        \mathcal{P} \left(\max_{i \in [1,K]}|\mathcal{E}_{\mathbb{D}} (\theta_i) - \hat{\mathcal{E}}_{D} (\theta_i)| \geq \epsilon \right) & \leq \sum_{i=1}^{K} \mathcal{P} \left(|\mathcal{E}_{\mathbb{D}} (\theta_i) - \hat{\mathcal{E}}_{D} (\theta_i)| \geq \epsilon \right) \\
        & \leq \sum_{i=1}^{K} 2  m_{{\Theta_i}}(N) \exp(-2N\epsilon^{2}) ,
    \end{split}
\end{align}
where $m_{{\Theta_i}}(N)$ is the amount of all possible prediction results for $N$ samples, which implies the model complexity in the parameter space $\Theta_i$. We set $m_{{\Theta_i}}(N) = \frac{1}{2} \left(\frac{ N}{d_i}\right)^{d_i}$ in our model, and assume a confidence bound $\epsilon_i = \sqrt{\frac{{\textit d_i} [\ln (\textit N /\textit d_i) ] + \ln (K/\delta)}{2 \textit N} }$, and $\epsilon = \max_{i \in [1,K]} \epsilon_i$. Then we get
\begin{align}
    \begin{split}
        \mathcal{P} \left(\max_{i \in [1,K]}|\mathcal{E}_{\mathbb{D}} (\theta_i) - \hat{\mathcal{E}}_{D} (\theta_i)| \geq \epsilon \right) &  \leq \sum_{i=1}^{K} 2  m_{{\Theta_i}}(N) \exp(-2N\epsilon^{2}) \\
        &  = \sum_{i=1}^{K}  \left(\frac{ N}{d_i}\right)^{d_i}\exp(-2N\epsilon^{2}) \\
        & \leq  \sum_{i=1}^{K}  \left(\frac{ N}{d_i}\right)^{d_i}\exp(-2N {\epsilon_i}^{2}) \\
        & = \sum_{i=1}^{K}  \frac{\delta}{K} = \delta.
    \end{split}
\end{align}

Hence, the inequality $|\mathcal{E}_{\mathbb{D}} (\theta) - \hat{\mathcal{E}}_{D} (\theta) |  \leq \epsilon $ holds with probability at least $1-\delta$. Further, based on the fact that $\hat{\mathcal{E}}_{D}^b (\theta) \geq \hat{\mathcal{E}}_{D} (\theta)$, we have
\begin{align}
    |\mathcal{E}_{\mathbb{D}} (\theta) - \hat{\mathcal{E}}_{D}^b (\theta) | \leq |\mathcal{E}_{\mathbb{D}} (\theta) - \hat{\mathcal{E}}_{D} (\theta) | \leq \epsilon .
\end{align}
It completes the proof.


\noindent \textbf{Proof of Proposition~\ref{coscl}} 
Let $\{\Theta_i \in  \mathbb{R}^{r}\}_{i=1}^{K}$ be a set of $K$ parameter spaces ($K>1$ in general), 
$d_i$ be a VC dimension of $\Theta_i$, and $\Theta = \cup_{i=1}^{K}\Theta_i$ with VC dimension $d$.
Let $\hat \theta_{1:t}^{b}$ denote the optimal solution of the continually learned $1:t$ tasks by
robust empirical risk minimization over the new task, i.e., $\hat \theta_{1:t}^{b} = \arg \min_{\theta \in \Theta} \hat{\mathcal{E}}_{D_t}^b (\theta)$, where $\Theta$ denotes a cover of a parameter space with VC dimension $d$.
Likewise, let $\hat \theta_{t}^{b}$ be the optimal solution by robust empirical risk minimization over the distribution $\mathbb{D}_t$ only, and $\hat \theta_{1:t-1}^{b}$ over the set of distribution $\mathbb{D}_1,\cdots,\mathbb{D}_{t-1}$. That is, $\hat \theta_{t}^{b} = \arg \min_{\theta} \hat{\mathcal{E}}_{D_t}^b (\theta)$ and $\hat \theta_{1:t-1}^{b} = \arg \min_{\theta} \hat{\mathcal{E}}_{D_{1:t-1}}^b (\theta)$.

Then, let $ \theta_{t}$ be the optimal solution over the distribution $\mathbb{D}_t$ only, i.e., $\theta_{t} = \arg \min_{\theta} {\mathcal{E}}_{\mathbb{D}_t} (\theta)$. From Lemma~\ref{lemma2} and Proposition~\ref{gapbound}, the following inequality holds with probability at least $1-\frac{\delta}{2}$,
\begin{align} \label{pro3-Dt-1}
\begin{split}
    |\mathcal{E}_{\mathbb{D}_{t}} (\theta_{1:t-1}) - \hat{\mathcal{E}}_{D_{t}} (\theta_{1:t-1})|  \leq \sqrt{\frac{d\ln (N_{t}/d) + \ln (2/\delta)}{N_{t}} },
\end{split}
\end{align}
where $N_{1:t-1}=\sum_{k=1}^{t-1} N_k$ is the total number of training samples over all $t-1$ old tasks. Then, we have
\begin{align} \label{pro3-dd}
\begin{split}
    \min_{\theta \in \Theta} \hat{\mathcal{E}}_{D_{1:t-1}} (\theta) 
    \leq \min_{\theta \in \Theta} \mathcal{E}_{\mathbb{D}_{t}} (\theta )  + \frac{1}{2(t-1)}\sum_{k=1}^{t-1} \rm{Div}(\mathbb{D}_\textit{k}, \mathbb{D}_{\textit{t}})   + \sqrt{\frac{\textit{d}\ln (\textit{N}_{\textit{t}}/\textit{d}) + \ln (2/\delta)}{\textit{N}_{\textit{t}}} }.
    \end{split} 
\end{align}

From Proposition~\ref{errorbound} and Lemma~\ref{pro3-lemma}, the following inequality holds with probability at least $1-\frac{\delta}{2}$,
\begin{align} \label{pro3-edt} 
\begin{split}
     \mathcal{E}_{\mathbb{D}_t} (\hat \theta_{1:t-1}^{b}) <  \hat{\mathcal{E}}_{D_{1:t-1}}^b (\hat \theta_{1:t-1}^{b})& + \frac{1}{2(t-1)}\sum_{k=1}^{t-1} \rm{Div}(\mathbb{D}_\textit{k}, \mathbb{D}_\textit{t}) \\
     & + \max_{i \in [1,K]} \sqrt{\frac{d_i\ln (N_{1:\textit t-1}/d_i) + \ln (2K/\delta)}{2N_{1:\textit t-1}} }.
     \end{split}
\end{align}

Combining Eqn.~\ref{pro3-dd} and Eqn.~\ref{pro3-edt}, we get
\begin{align} \label{pro3-part1}
    \begin{split}
      & \mathcal{E}_{\mathbb{D}_{t}} (\hat \theta_{1:t}^{b}) - \min_{\theta \in \Theta} \mathcal{E}_{\mathbb{D}_{t}} (\theta )  \leq \mathcal{E}_{\mathbb{D}_{t}} (\hat \theta_{1:t-1}^{b}) - \min_{\theta \in \Theta} \mathcal{E}_{\mathbb{D}_{t}} (\theta ) \\
       &  \leq \hat{\mathcal{E}}_{D_{1:t-1}}^b (\hat \theta_{1:t-1}^{b}) - \min_{\theta \in \Theta} \hat{\mathcal{E}}_{D_{1:t-1}} (\theta)  +\frac{1}{t-1}\sum_{k=1}^{t-1} \rm{Div}(\mathbb{D}_\textit{k}, \mathbb{D}_\textit{t})\\
       & + \max_{i \in [1,K]} \sqrt{\frac{d_i\ln (N_{1:\textit t-1}/d_i) + \ln (2K/\delta)}{2N_{1:\textit t-1}} }+\sqrt{\frac{{\textit d}\ln (\textit N_{1:\textit t-1}/\textit d) + \ln (2/\delta)}{\textit N_{1:\textit t-1}} } \\
       & =\min_{\theta \in \Theta} \hat{\mathcal{E}}_{D_{1:t-1}}^b (\theta) - \min_{\theta \in \Theta} \hat{\mathcal{E}}_{D_{1:t-1}} (\theta)  +\frac{1}{t-1}\sum_{k=1}^{t-1} \rm{Div}(\mathbb{D}_\textit{k}, \mathbb{D}_\textit{t}) \\
       & + \max_{i \in [1,K]} \sqrt{\frac{d_i\ln (N_{1:\textit t-1}/d_i) + \ln (2K/\delta)}{2N_{1:\textit t-1}} } + \sqrt{\frac{{\textit d}\ln (\textit N_{1:\textit t-1}/\textit d) + \ln (2/\delta)}{\textit N_{1:\textit t-1}} }. 
    \end{split}
\end{align}
This completes the first part of Proposition~\ref{coscl}.

Similarly, let $ \theta_{1:t}$ be the optimal solution over the distribution $\mathbb{D}_{1:t-1}$ only, i.e., $\theta_{1:t-1} = \arg \min_{\theta} {\mathcal{E}}_{\mathbb{D}_{1:t-1}} (\theta)$. From Lemma~\ref{lemma2} and Proposition~\ref{gapbound}, the following inequality holds with probability at least $1-\frac{\delta}{2}$,
\begin{align} \label{pro3-Dt-2}
    |\mathcal{E}_{\mathbb{D}_{t}} (\theta_{1:t-1}) - \hat{\mathcal{E}}_{D_{t}} (\theta_{1:t-1})| \leq \sqrt{\frac{d\ln (N_{t}/d) + \ln (2/\delta)}{N_{t}} },
\end{align}
where $N_t$ is the number of training samples in the distribution $\mathbb{D}_{t}$. Then, we have 
\begin{small}
\begin{align} \label{pro3-dd-2}
\begin{split}
    \min_{\theta \in \Theta} \hat{\mathcal{E}}_{D_{t}} (\theta) 
    \leq \min_{\theta \in \Theta} \mathcal{E}_{\mathbb{D}_{1:t-1}} (\theta )  + \frac{1}{2(t-1)}\sum_{k=1}^{t-1} \rm{Div}(\mathbb{D}_\textit{t}, \mathbb{D}_{\textit{k}})   + \sqrt{\frac{\textit{d}\ln (\textit{N}_{\textit{t}}/\textit{d}) + \ln (2/\delta)}{\textit{N}_{\textit{t}}} }.
    \end{split} 
\end{align}
\end{small}
From Proposition~\ref{errorbound} and Lemma~\ref{pro3-lemma}, the following inequality holds with probability at least $1-\frac{\delta}{2}$,
\begin{small}
\begin{align} \label{pro3-edt-2} 
     \mathcal{E}_{\mathbb{D}_{1:t-1}} (\hat \theta_{t}^{b}) <  \hat{\mathcal{E}}_{D_{t}}^b (\hat \theta_{t}^{b}) + \frac{1}{2(t-1)}\sum_{k=1}^{t-1} \rm{Div}(\mathbb{D}_\textit{t}, \mathbb{D}_\textit{k}) + 
        \max_{i \in [1,K]} \sqrt{\frac{\textit{d}_\textit{i}\ln (\textit N_{\textit t}/\textit{d}_\textit{i}) + \ln (2\textit{K}/\delta)}{2\textit N_{\textit t}} }.
\end{align}
\end{small}

Combining Eqn.~\ref{pro3-dd-2} and Eqn.~\ref{pro3-edt-2}, we get
\begin{align}
    \begin{split} \label{pro3-part2}
      & \mathcal{E}_{\mathbb{D}_{1:t-1}} (\hat \theta_{1:t}^{b}) - \min_{\theta \in \Theta} \mathcal{E}_{\mathbb{D}_{1:t-1}} (\theta )  \leq  \mathcal{E}_{\mathbb{D}_{1:t-1}} (\hat \theta_{t}^{b}) - \min_{\theta \in \Theta} \mathcal{E}_{\mathbb{D}_{1:t-1}} (\theta )  \\
       &  \leq \hat{\mathcal{E}}_{D_{t}}^b (\hat \theta_{t}^{b})  -\min_{\theta \in \Theta} \hat{\mathcal{E}}_{D_{t}} (\theta)  +\frac{1}{t-1}\sum_{k=1}^{t-1} \rm{Div}(\mathbb{D}_\textit{t}, \mathbb{D}_\textit{k}) \\
       & + \max_{i \in [1,K]} \sqrt{\frac{d_i\ln (\textit N_{\textit t}/d_i) + \ln (2K/\delta)}{2\textit N_{\textit t}} }
       + \sqrt{\frac{{\textit d}\ln (\textit N_{\textit t}/\textit d) + \ln (1/\delta)}{\textit N_{\textit t}} } \\
       & = \min_{\theta \in \Theta} \hat{\mathcal{E}}_{D_{t}}^b (\theta) - \min_{\theta \in \Theta} \hat{\mathcal{E}}_{D_{t}} (\theta)  +\frac{1}{t-1}\sum_{k=1}^{t-1} \rm{Div}(\mathbb{D}_\textit{t}, \mathbb{D}_\textit{k}) \\
       & + \max_{i \in [1,K]} \sqrt{\frac{d_i\ln (\textit N_{\textit t}/d_i) + \ln (2K/\delta)}{2\textit N_{\textit t}} } + \sqrt{\frac{{\textit d}\ln (\textit N_{\textit t}/\textit d) + \ln (1/\delta)}{\textit N_{\textit t}} } .
    \end{split}
\end{align}
This completes the second part of Proposition~\ref{coscl}.


\textbf{Discrepancy between task distributions:} 
Below are three important lemmas to prove how cooperating multiple continual learners can optimize the discrepancy between task distributions, which is measured by $\mathcal{H}$-divergence.

\begin{lemma}\label{div}
(Based on Theorem 3.4 of \cite{kifer2004detecting} and Lemma 1 of \cite{ben2010theory})
Let $\Theta$ be a cover of a parameter space with VC dimension $d$.
If $T$ and $S$ are samples of size $N$ from two distributions $\mathbb{T}$ and $\mathbb{S}$, respectively, and $\hat {\rm{Div}}(T,S)$ is the empirical $\mathcal{H}$-divergence between samples, then for any $\delta \in (0,1)$, with probability at least $1-\delta$,
\begin{align}
    \rm{Div}(\mathbb{T}, \mathbb{S}) \leq \hat {\rm{Div}}(\textit{T},\textit{S}) + 4\sqrt{\frac{\textit{d} \ln(2\textit{N}) + \ln({2}/{\delta})}{\textit{N}}}.
\end{align}
\end{lemma}

\begin{lemma} \label{emDIV}
(Based on Lemma 2 of \cite{ben2010theory})
Let $T$ and $S$ be samples of size $N$ from two distributions $\mathbb{T}$ and $\mathbb{S}$, respectively.
Then the empirical $\mathcal{H}$-divergence between samples, i.e., $\hat {\rm{Div}}(T,S)$ can be computed by finding a classifier which attempts to separate one distribution from the other.
That is,
\begin{align}
    \hat {\rm{Div}}(T,S) = 2\left(1- \frac{1}{N} \min_{\theta \in \Theta}\left[  \sum_{x:p_{\theta}(x)=0} I[x \in \mathbb{S} ]+  \sum_{x:p_{\theta}(x)=1} I[x \in \mathbb{T} ] \right]\right),
\end{align}
where $I[x \in \mathbb{S} ]$ is the binary indicator variable which is 1 when the input $x \in \mathbb{S}$, and 0 when $x \in \mathbb{T}$. $p_{\theta}(\cdot)$ is the learned prediction function.
\end{lemma}
Of note, Lemma~\ref{emDIV} implies we first find a solution in parameter space which has minimum error for the binary problem of distinguishing source from target distributions. By cooperating $K$ parameter spaces, i.e., $\Theta = \cup_{i=1}^{K}\Theta_i$, we can improve classification errors so as to decrease $\mathcal{H}$-divergence. 

\begin{lemma} \label{ours-div}
Let $\{\Theta_i \in  \mathbb{R}^{r}\}_{i=1}^{K}$ be a set of $K$ parameter spaces ($K>1$ in general), 
$d_i$ be a VC dimension of $\Theta_i$, and $\Theta = \cup_{i=1}^{K}\Theta_i$ with VC dimension $d$.
If $T$ and $S$ are samples of size $N$ from two distributions $\mathbb{T}$ and $\mathbb{S}$, respectively,
and $\hat {\rm{Div}}(T,S)$ is the empirical $\mathcal{H}$-divergence between samples,
then in the parameter space $\Theta$, for any $\delta \in (0,1)$, with probability at least $1-\delta$,
\begin{align}
    \rm{Div}(\mathbb{T}, \mathbb{S}) \leq \hat {\rm{Div}}(\textit{T},\textit{S}) + \max_{i \in [1,\textit{K}]} 4\sqrt{\frac{\textit{d}_i \ln(2\textit{N}) + \ln({2\textit{K}}/{\delta})}{2\textit{N}}}.
\end{align}
\end{lemma}
\noindent\emph{Proof.}
For two distributions $\mathbb{T}$ and $\mathbb{S}$, we have
\begin{align}
    \begin{split}
        &\mathcal{P} \left(\max_{i \in [1,K]}|\rm{Div}_{\Theta_i}(\mathbb{T}, \mathbb{S}) - \hat {\rm{Div}}_{\Theta_i}(\textit{T},\textit{S})| \geq \epsilon \right) \\
        & \leq \sum_{i=1}^{K} \mathcal{P} \left(|\rm{Div}_{\Theta_i}(\mathbb{T}, \mathbb{S}) - \hat {\rm{Div}}_{\Theta_i}(\textit{T},\textit{S}) | \geq \epsilon \right)  \leq \sum_{i=1}^{K} 2  m_{{\Theta_i}}(N) \exp(-2N\epsilon^{2}) ,
    \end{split}
\end{align}
where $m_{{\Theta_i}}(N)$ is the amount of all possible predictions for $N$ samples, which implies the model complexity in the parameter space $\Theta_i$. We set $m_{{\Theta_i}}(N) = 16 \left(2N\right)^{d_i}$ in our model, and assume a confidence bound $\epsilon_i = 4 \sqrt{\frac{d_i \ln(2N) + \ln({2K}/{\delta})}{2N}}$, and $\epsilon = \max_{i \in [1,K]} \epsilon_i$. Then we get
\begin{align}
    \begin{split}
        \mathcal{P} \left(\max_{i \in [1,K]}|\rm{Div}_{\Theta_i}(\mathbb{T}, \mathbb{S}) - \hat {\rm{Div}}_{\Theta_i}(\textit{T},\textit{S})| \geq \epsilon \right) &  \leq \sum_{i=1}^{K} 2  m_{{\Theta_i}}(N) \exp(-2N\epsilon^{2}) \\
        &  = \sum_{i=1}^{K}  32\left(2N\right)^{d_i}\exp(-2N\epsilon^{2}) \\
        & \leq  \sum_{i=1}^{K} 32 \left(2N\right)^{d_i}\exp(-2N {\epsilon_i}^{2}) \\
        & = \sum_{i=1}^{K}  \frac{\delta}{K} = \delta.
    \end{split}
\end{align}
Hence, the inequality $|\rm{Div}_{\Theta_i}(\mathbb{T}, \mathbb{S}) - \hat {\rm{Div}}_{\Theta_i}(\textit{T},\textit{S}) |  \leq \epsilon $ holds with probability at least $1-\delta$. 
It completes the proof.

Comparing Lemma~\ref{div} and Lemma~\ref{ours-div}, it can be found that by cooperating $K$ parameter spaces, our proposal can mitigate the discrepancy between tasks, i.e., $\rm{Div}(\mathbb{T}, \mathbb{S}) $, by decreasing the empirical $\mathcal{H}$-divergence (i.e., $\hat {\rm{Div}}_{\Theta_i}(\textit{T},\textit{S})$) and another factor.

\clearpage
\section{Experiment Details}

\subsection{Implementation}
We follow the implementation of \cite{jung2020continual,cha2020cpr,wang2021afec} for supervised continual learning. For CIFAR-100-SC and CIFAR-100-RS, we use an Adam optimizer of initial learning rate 0.001 and train all methods with batch size of 256 for 100 epochs. For CUB-200-2011 and Tiny-ImageNet, we use a SGD optimizer of initial learning rate 0.005 and momentum 0.9, and train all methods with batch size of 64 for 40 epochs. 

We follow the implementation of \cite{madaan2021rethinking} for unsupervised continual learning on CIFAR-100-RS (which is called Split CIFAR-100 in \cite{madaan2021rethinking}). We use a SGD optimizer of initial learning rate 0.03, momentum 0.9 and weight decay 5e-4, and train all methods with batch size of 256 for 200 epochs.

\subsection{Hyperparameter}

For CIFAR-100-SC, CIFAR-100-RS and CUB-200-2011, we adopt the same hyperparameters for the baselines used in \cite{wang2021afec}. While for other experiments (e.g., Tiny-ImageNet) and baselines (e.g., CPR \cite{cha2020cpr}), we make an extensive hyperparameter search to make the comparison as fair as possible. The hyerparameters for supervised continual learning are summarized in Table~\ref{table:hyperparameter_all}. 

\begin{table*}[h]
	\centering
     \vspace{-.2cm}
	\caption{Hyperparamters for supervised continual learning. \(^*\)\(\lambda\) is the same as the corresponding baseline approach.} 
    \vspace{-.1cm}
	\smallskip
	\resizebox{0.90\textwidth}{!}{ 
	\begin{tabular}{ccccccc}
		\specialrule{0.01em}{1.2pt}{1.5pt}
       Methods & CIFAR-100-SC & CIFAR-100-RS & CUB-200-2011 & Tiny-ImageNet \\
       \specialrule{0.01em}{1.2pt}{1.7pt}
       AGS-CL \cite{jung2020continual} & \(\lambda\)(3200), \(\mu\)(10), \(\rho\)(0.3)  & \(\lambda\)(1600), \(\mu\)(10), \(\rho\)(0.3) &-- &-- \\
        HAT \cite{serra2018overcoming} &  \(c\)(500), \(\rm{smax}\)(200) &  \(c\)(500), \(\rm{smax}\)(200)  &-- &-- \\
       EWC \cite{kirkpatrick2017overcoming}&\(\lambda\)(40000) &\(\lambda\)(10000) & \(\lambda\)(1) & \(\lambda\)(80) \\
       MAS \cite{aljundi2018memory}&\(\lambda\)(16) &\(\lambda\)(4) &\(\lambda\)(0.01) &\(\lambda\)(0.1) \\
       SI \cite{zenke2017continual}&\(\lambda\)(8) &\(\lambda\)(10) &\(\lambda\)(6) &\(\lambda\)(0.8) \\
       RWALK \cite{chaudhry2018riemannian} &\(\lambda\)(128) &\(\lambda\)(6) &\(\lambda\)(48) &\(\lambda\)(5) \\
       P\&C \cite{schwarz2018progress}& \(\lambda\)(40000) & \(\lambda\)(20000) & \(\lambda\)(1) & \(\lambda\)(80) \\
       \(^*\)AFEC \cite{wang2021afec}& \(\lambda_e\)(1) & \(\lambda_e\)(1) & \(\lambda_e\)(0.001) & \(\lambda_e\)(0.1) \\
       \(^*\)CPR \cite{cha2020cpr}&\(\beta\)(1.5) &\(\beta\)(1.5) &\(\beta\)(1) &\(\beta\)(0.6) \\
       \(^*\)CoSCL (Ours)& \(\gamma\)(0.02), \(s\)(100) & \(\gamma\)(0.02), \(s\)(100) &\(\gamma\)(0.0001), \(s\)(100) &\(\gamma\)(0.001), \(s\)(100) \\
      \specialrule{0.01em}{1.2pt}{1.7pt}
	\end{tabular}
	}
	\label{table:hyperparameter_all}
	\vspace{-.5cm}
\end{table*}

\subsection{Architecture}
The network architectures used for the main experiments are detailed in Table~\ref{table:architecture_6cnn}, \ref{table:architecture_alexnet} (the output head is not included). 

\subsection{Evaluation Metric}
We use three metrics to evaluate the performance of continual learning, including averaged accuracy (AAC), forward transfer (FWT) and backward transfer (BWT) \cite{lopez2017gradient}:

\begin{equation}
    \rm{AAC} = \frac{1}{T} \sum_{i=1}^{T} A_{T,i},
\end{equation}
\begin{equation}
    \rm{FWT} = \frac{1}{T-1} \sum_{i=2}^{T} A_{i-1,i} - \hat{A}_{i},
\end{equation}
\begin{equation}
    \rm{BWT} = \frac{1}{T-1} \sum_{i=1}^{T-1} A_{T,i} - A_{i,i},
\end{equation}
where \(\rm{A}_{t,i}\) is the test accuracy of task \(i\) after incrementally learning task \(t\), and \(\rm{\hat{A}}_{i}\) is the test accuracy of each task \(i\) learned from random initialization. Averaged accuracy (ACC) is the averaged performance of all the tasks ever seen. Forward transfer (FWT) evaluates the averaged influence of remembering the old tasks to each new task. Backward transfer (BWT) evaluates the averaged influence of learning each new task to the old tasks. 

\section{Additional Results}

\subsection{Diversity of Expertise across Tasks}

\begin{figure}[ht]
	\centering
	\includegraphics[width=0.85\columnwidth]{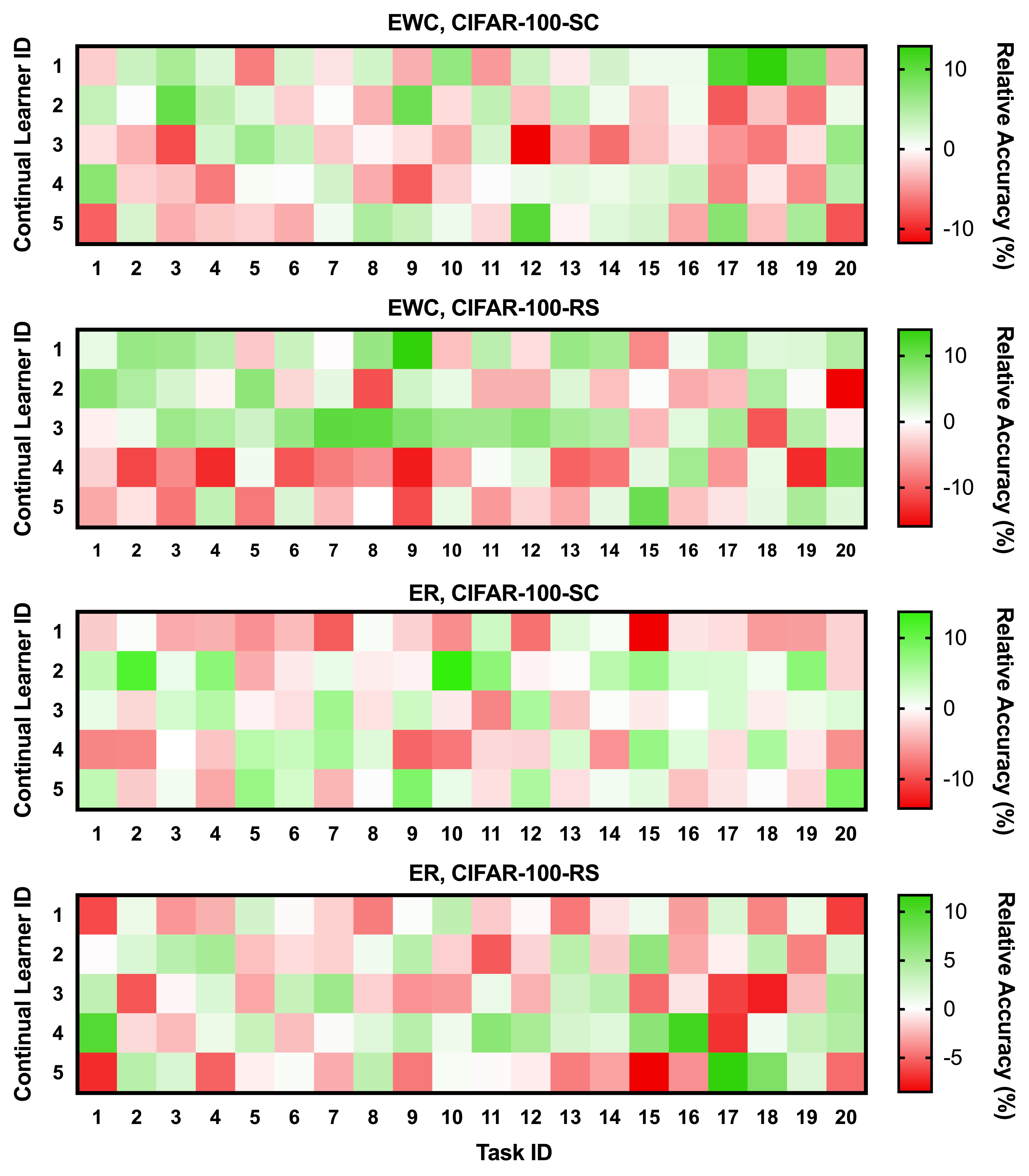} 
    \vspace{-.2cm}
	\caption{Diversity of expertise across tasks. Here we use EWC \cite{kirkpatrick2017overcoming} or Experience Replay (ER) \cite{riemer2018learning} as the default continual learning method. The relative accuracy for each task is calculated by subtracting the performance of each learner from the averaged performance of all learners.}
	\label{fig:heatmap_all}
\end{figure}

To evaluate the diversity of expertise across tasks, we use the feature representations of each continual learner to make predictions with the shared output head, and calculate the relative accuracy. As shown in Fig.~\ref{fig:heatmap_all}, the solution learned by each continual learner varies significantly across tasks and complement with each other.

\subsection{Discrepancy between Task Distributions}


To empirically approximate the $\mathcal{H}$-divergence between tasks in feature space, we train a discriminator with a fully-connected layer to distinguish whether the features of input images belong to a task or not \cite{long2015learning}. Specifically, the discriminator is trained with the features of training data and the binary cross-entropy loss. We use Adam optimizer and initial learning rate 0.0001 with batch size of 256 for 10 epochs. Then we evaluate the $\mathcal{H}$-divergence between tasks with the features of test data, where a larger discrimination loss indicates a smaller $\mathcal{H}$-divergence. Since the discrimination becomes increasingly harder as more tasks are introduced, from the tenth task we start to observe significant differences between all the baselines. The proposed feature ensemble (FE) and ensemble cooperation (EC) can largely decrease the discrepancy between tasks, while the task-adaptive gates (TG) have a moderate effect.

\subsection{Results of ResNet}
In addition to regular CNN architectures, our method is also applicable to other architectures such as ResNet. We use a WideResNet-28-2 architecture to perform the task incremental learning experiments on CIFAR-100-RS, following a widely-used implementation code \cite{Hsu18_EvalCL}. CoSCL (5 learners with accordingly-adjusted width) can improve the performance from 69.52\% to 73.26\% for EWC and from 62.23\% to 68.69\% for MAS.


\begin{table*}[th]
	\centering
	\caption{Network architecture for CIFAR-100-SC and CIFAR-100-RS. We set \(nc=32\) for SCL (\#Param=837K) while \(nc=8\) for 5 learners in CoSCL (\#Param=773K).}
	\smallskip
	\resizebox{0.70\textwidth}{!}{ 
	\begin{tabular}{ccccccc}
		\specialrule{0.01em}{1.2pt}{1.5pt}
       Layer & Channel & Kernel & Stride & Padding & Dropout\\
      \specialrule{0.01em}{1.2pt}{1.7pt}
       Input & 3 &  & & & \\
       Conv 1 & \(nc\) & 3$\times$3  & 1 & 1 &  \\
       Conv 2 & \(nc\) &  3$\times$3  &1  &1 &   \\
       MaxPool &   &   &2 &0 &0.25  \\
       Conv 3 & \(2nc\)  & 3$\times$3  &1  &1 &    \\
       Conv 4 & \(2nc\)  & 3$\times$3  &1  &1  &   \\
       MaxPool &  &  &2 &0 &0.25   \\
       Conv 5 & \(4nc\) & 3$\times$3 &1  & 1 &   \\
       Conv 6 & \(4nc\) & 3$\times$3 &1 & 1  &    \\
       MaxPool&  &  &2 &1 &0.25   \\
       Dense 1& 256 &  &  &  &   \\
      \specialrule{0.01em}{1.2pt}{1.7pt}
	\end{tabular}
	}
	\label{table:architecture_6cnn}
	\vspace{-.6cm}
\end{table*}

\begin{table*}[th]
	\centering
	\caption{Network architecture for CUB-200-2011 and Tiny-ImageNet. We set \(nc=64\) for SCL (\#Param=57.8M) while \(nc=34\) for 5 learners in CoSCL (\#Param=57.2M). } 
	\smallskip
	\resizebox{0.70\textwidth}{!}{ 
	\begin{tabular}{ccccccc}
		\specialrule{0.01em}{1.2pt}{1.5pt}
       Layer & Channel & Kernel & Stride & Padding & Dropout\\
      \specialrule{0.01em}{1.2pt}{1.7pt}
       Input & 3 &  & & &  \\
       Conv 1 & \(nc\) & 11$\times$11  & 4 & 2 &  \\
       MaxPool &  & 3$\times$3  &2 &0 & 0 \\
       Conv 2 & \(3nc\) &  5$\times$5  &1  &2   &     \\
       MaxPool &  & 3$\times$3  &2 &0 & 0  \\
       Conv 3 & \(6nc\)  & 3$\times$3  &1  &1 &     \\
       Conv 4 & \(4nc\)  & 3$\times$3  &1  &1  &   \\
        Conv 5 & \(4nc\) & 3$\times$3 &1  & 1  &   \\
        MaxPool &  & 3$\times$3  &2 &0 & 0  \\
       Dense 1& \(64nc\) &  &  &   &0.5  \\
       Dense 2& \(64nc\) &  &  &   &0.5 \\
      \specialrule{0.01em}{1.2pt}{1.7pt}
	\end{tabular}
	}
	\label{table:architecture_alexnet}
	\vspace{-.4cm}
\end{table*}

\begin{figure}[ht]
	\centering
	\includegraphics[width=0.60\columnwidth]{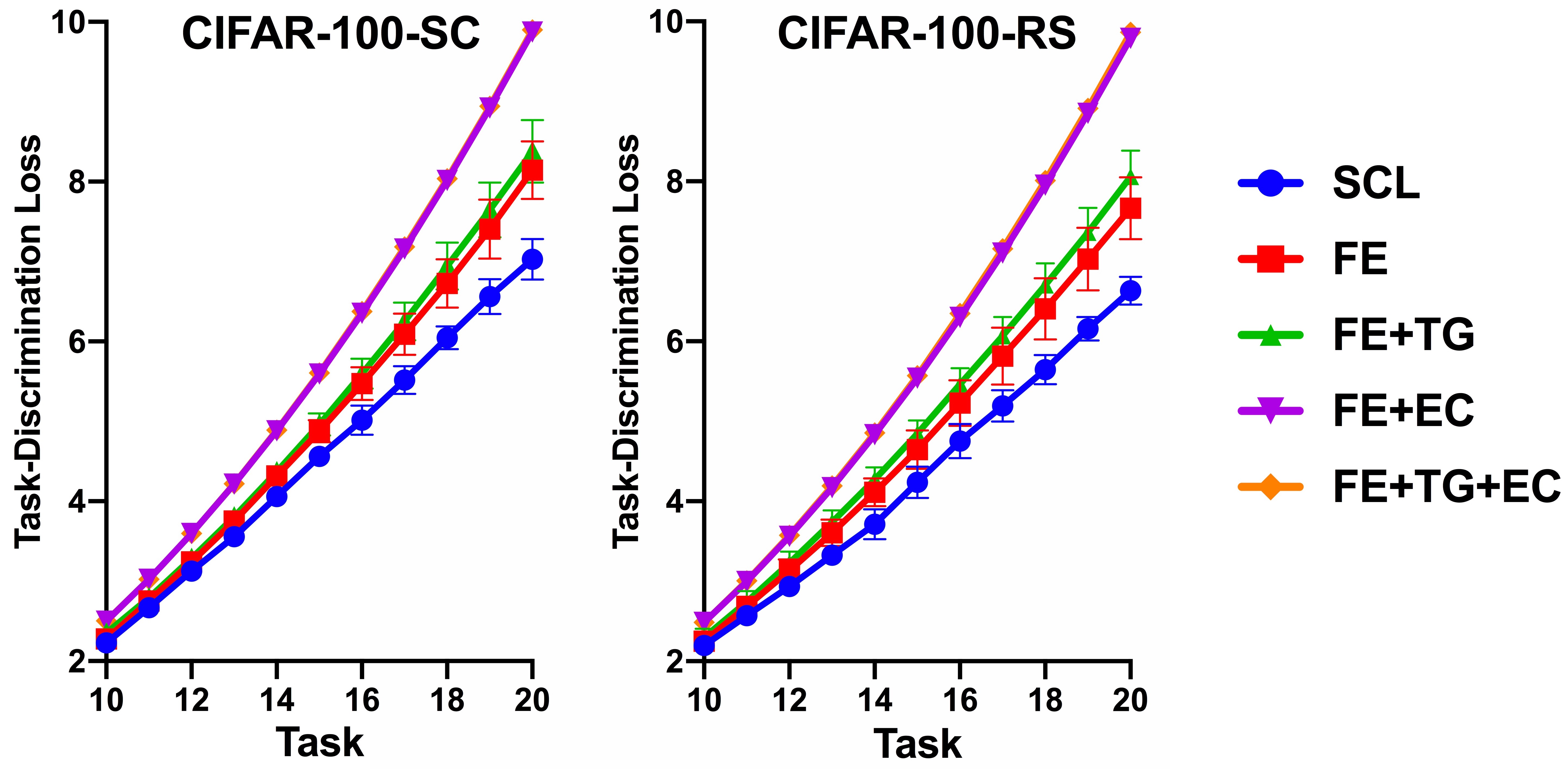} 
    \vspace{-.2cm}
	\caption{Task-discrimination loss in feature space. We plot all baselines from the tenth task, where significant differences start to arise. Larger loss indicates a smaller $\mathcal{H}$-divergence. SCL: single continual learner; FE: feature ensemble; TG: task-adaptive gates; EC: ensemble cooperation loss.}
	\label{fig:h_divergence_all}
    \vspace{-.2cm}
\end{figure}

\clearpage

\end{document}